\documentclass[twoside]{article}

\usepackage[accepted]{aistats2020}
\usepackage[round]{natbib}

\usepackage{amsthm}
\usepackage{amsmath}
\usepackage{amssymb}
\usepackage{graphicx}
\usepackage{color}
\usepackage{ifpdf}
\usepackage{url}
\usepackage{hyperref}
\usepackage{algorithm}
\usepackage{algorithmicx}
\usepackage[usenames,dvipsnames]{xcolor}
\usepackage{paralist}
\usepackage{pifont}
\usepackage{multirow}

\usepackage[ruled,vlined, linesnumbered, algo2e]{algorithm2e}
\usepackage{booktabs} 

\newtheorem{theorem}{Theorem}[section]
\newtheorem{corollary}{Corollary}[section]
\newtheorem{lemma}{Lemma}[section]
\newtheorem{assumption}{Assumption}[section]
\newtheorem{remark}{Remark}[section]

\newcommand{\R}{\mathbb{R}}

\newcommand{\abs}[1]{\left\vert#1\right\vert}

\newcommand{\set}[1]{\left\{#1\right\}}

\newcommand{\norm}[1]{\left\Vert#1\right\Vert}
\newcommand{\norms}[1]{\Vert#1\Vert}

\newcommand{\Eproof}{\hfill $\square$}
\newcommand{\prox}{\mathrm{prox}}

\newcommand{\argmin}{\mathrm{arg}\!\displaystyle\min}

\newcommand{\zero}[1]{{\boldsymbol{0}}}

\newcommand{\Ac}{\mathcal{A}}

\newcommand{\Gc}{\mathcal{G}}

\newcommand{\Bc}{\mathcal{B}}

\newcommand{\Sc}{\mathcal{S}}

\newcommand{\Tc}{\mathcal{T}}
\newcommand{\Rc}{\mathcal{R}}

\newcommand{\Fc}{\mathcal{F}}

\newcommand{\Pc}{\mathcal{P}}

\newcommand{\Exp}[1]{\mathbb{E}\left[#1\right]}
\newcommand{\Exps}[2]{\mathbb{E}_{#1}\left[#2\right]}

\newcommand{\Var}[1]{\mathrm{Var}\left[#1\right]}

\newcommand{\BigO}[1]{\mathcal{O}\left(#1\right)}

\newcommand{\dsum}{{\displaystyle\sum}}

\newcommand{\cmark}{\ding{51}}%
\newcommand{\xmark}{\ding{55}}%
\newcommand{\done}{\rlap{$\square$}{\raisebox{2pt}{\large\hspace{1pt}\cmark}}%
\hspace{-2.5pt}}
\newcommand{\wontfix}{\rlap{$\square$}{\large\hspace{1pt}\xmark}}

\newcommand{\beforesec}{\vspace{0ex}}
\newcommand{\aftersec}{\vspace{0ex}}

\newcommand{\beforepara}{\vspace{-2ex}}

\begin{document}

%

%
\runningauthor{Pham, Nguyen, Phan, Nguyen, van Dijk, Tran-Dinh}

\twocolumn[

\aistatstitle{A Hybrid Stochastic Policy Gradient Algorithm for Reinforcement Learning}

\aistatsauthor{$^{*}$Nhan H. Pham \And $^{\dagger}$Lam M. Nguyen \And  $^{\dagger}$Dzung T. Phan \And $^{\ddagger}$Phuong Ha Nguyen}
\vspace{1ex}
\aistatsauthor{$^{\ddagger}{^{\S}}$Marten van Dijk \And $^{*}$Quoc Tran-Dinh}

\vspace{1ex}
%

\aistatsaddress{ $^{*}$UNC Chapel Hill \And $^{\dagger}$IBM Research \And $^{\ddagger}$UConn \And ${^{\S}}$CWI Amsterdam}

%

]

\begin{abstract}
\vspace{-2ex}
We propose a novel hybrid stochastic policy gradient estimator by combining an unbiased policy gradient estimator, the REINFORCE estimator, with another biased one, an adapted SARAH estimator for policy optimization.
The hybrid policy gradient estimator is shown to be biased, but has variance reduced property.
Using this estimator, we develop a new Proximal Hybrid Stochastic Policy Gradient Algorithm (ProxHSPGA) to solve a composite policy optimization problem that allows us to handle constraints or regularizers on the policy parameters. 
We first propose a single-looped algorithm then introduce a more practical restarting variant.
We prove that both algorithms can achieve the best-known trajectory complexity $\BigO{\varepsilon^{-3}}$ to attain a first-order stationary point for the composite problem which is better than existing REINFORCE/GPOMDP $\BigO{\varepsilon^{-4}}$ and SVRPG $\BigO{\varepsilon^{-10/3}}$ in the non-composite setting.
We evaluate the performance of our algorithm on several well-known examples in reinforcement learning. 
Numerical results show that our algorithm outperforms two existing methods on these examples.
Moreover, the composite settings indeed have some advantages compared to the non-composite ones on certain problems.
\end{abstract}

\section{Introduction}\label{sec:intro}
Recently, research on reinforcement learning (RL) \citep{Sutton2018}, an area of machine learning to learn how to make a series of decisions while interacting with the underlying environment, has been immensely active.
Unlike supervised learning, reinforcement learning agents often have limited or no knowledge about the environment and the rewards of taking certain actions might not be immediately observed, making these problems more challenging to solve.
Over the past decade, there has been a large number of research works developing and using reinforcement learning to solve emerging problems. 
Notable reinforcement learning agents include, but not limited to, AlphaGo and AlphaZero \citep{Silver2016,Silver2018}, OpenAIFive \citep{OpenAI_dota}, and AlphaStar \citep{alphastarblog}.

In modern RL tasks, the environment is often not known beforehand so the agent has to simultaneously learn the environment while making appropriate decisions.
One approach is to estimate the value function or the state-value function, e.g., Q-learning \citep{Watkins1992} and its variants such as Deep Q-learning (DQN) \citep{Mnih2013, Mnih2015}, Dueling DQN \citep{Wang2016}, and double Q-learning \citep{Hasselt2016}.

It has been observed that learning the state-value function is not efficient when the action space is large or even infinite.  
In that case, policy gradient methods learn the policy directly with a parameterized function.
\cite{Silver2014} presents a framework for deterministic policy gradient algorithms which can be estimated more efficiently than their stochastic counterparts whereas DDPG \citep{Lillicrap2016} adapts the idea of deep Q-learning into continuous action tasks in RL. 
TRPO \citep{Schulman2015} uses a constraint on the KL divergence between the new and old policies to improve the robustness of each update.
PPO \citep{Schulman2017} is an extension of TRPO which uses a clipped surrogate objective resulting a simpler implementation.
Other policy gradient methods utilize the actor-critic paradigm including ACER \citep{Wang2017}, A3C \citep{Mnih2016} and its synchronous variant A2C, ACKTR \citep{Wu2017}, and SAC \citep{Haarnoja2018}.

REINFORCE \citep{Williams1992} is perhaps one classical method closely related to our work here.
It uses an estimator of the policy gradient and applies a gradient ascent step to update the policy. 
Nevertheless, the REINFORCE estimator is known to have high variance leading to several weaknesses. 
Other improvements to reduce the variance such as adding baselines \citep{Sutton2018,Zhao2011}, discarding some rewards in the so-called GPOMDP estimator \citep{Baxter2001} were proposed. While REINFORCE estimator is an unbiased policy gradient estimator, GPOMDP is shown to be biased \citep{Baxter2001} making theoretical analysis harder.

The nature of REINFORCE algorithm appears to be closely related to stochastic gradient descent (SGD) \citep{Robbins1951} in stochastic nonconvex optimization. 
In particular, the standard SGD estimator is also known to often have fixed variance, which is often high. 
On the one hand, there are algorithms trying to reduce the oscillation \citep{Tieleman2012} or introduce momentums or adaptive updates \citep{Allen-Zhu2017, Allen-Zhu2018, Kingma2014} for SGD methods to accelerate performance. 
On the other hand, other researchers are searching for new gradient estimators.  
One approach is the SAGA estimator proposed by \cite{Defazio2014}.
Another well-known estimator is the SVRG estimator \citep{johnson2013accelerating} which has been intensively studied in recent works, e.g., in \cite{Allen-Zhu2016,Li2018,Reddi2016,Zhou2018StochasticNV}.
This estimator not only overcomes the storage issue of SAGA but also possesses variance reduced property, i.e., the variance of the estimator decreases over epochs. 
Methods based on SVRG estimators have recently been developed for reinforcement learning, e.g., SVRPG \citep{Papini2018a}. \cite{Xu2019} refines the analysis of SVRPG to achieve an improved trajectory complexity of $\BigO{\varepsilon^{-10/3}}$.
\cite{Shen2019d} also adopts the SVRG estimator into policy gradient and achieve the trajectory oracle complexity of $\BigO{\varepsilon^{-3}}$ with the use of a second-order estimator.

While SGD, SAGA, and SVRG estimators are unbiased, there have been algorithms developed based on a biased gradient estimator named SARAH \citep{Nguyen2017b}. 
Such algorithms include SARAH \citep{Nguyen2017,Nguyen2019_SARAH}, SPIDER \citep{Fang2018}, SpiderBoost \citep{Wang2018}, and ProxSARAH \citep{Pham2019ProxSARAH}.
Similar to SVRG, all these methods can potentially be extended to reinforcement learning.
A recent attempt is SARAPO \citep{yuan2019policy} which combines SARAH \citep{Nguyen2019_SARAH} with TRPO \citep{Schulman2015} algorithm but no theoretical guarantee is provided. 
\cite{Yang2019} propose Mirror Policy Optimization (MPO) algorithm which covers the classical policy gradient and the natural policy gradient as special cases. 
They also introduce a variance reduction variant, called VRMPO, which achieves $\BigO{\varepsilon^{-3}}$ trajectory complexity. 
Another notable work is SRVR-PG \citep{Xu2019SampleEP} where the policy gradient estimator is the adapted version of SARAH estimator for reinforcement learning. Note that \cite{Yang2019} and \cite{Xu2019SampleEP} achieve the same trajectory complexity of $\BigO{\varepsilon^{-3}}$ as ours. 
However, our algorithm is essentially different. 
\cite{Xu2019SampleEP} and \cite{Yang2019} use two different adaptation of the SARAH estimator for policy gradient. \cite{Xu2019SampleEP} uses the importance weight in their estimator to handle distribution shift while \cite{Yang2019} remove it as seen in \cite{Shen2019d}. 
Meanwhile, we introduce a new policy gradient estimator which can also be calculated recursively. 
The new estimator is fundamentally different from the other two since it combines the adapted SARAH estimator as in \cite{Xu2019SampleEP} with the classical REINFORCE estimator. 
In addition, our analysis shows that the best-known convergence rate and complexity can be achieved by our single-loop algorithm (Algorithm~\ref{alg:A1}) while SRVR-PG and VRMPO require double loops to achieve the same oracle complexity. Moreover, \cite{Xu2019SampleEP,Yang2019} do not consider the composite setting that considers both constraints and regularizers on the policy parameters as we do.

\begin{table}[hpt!]
\vspace{-2ex}
\newcommand{\cellb}[1]{{\!\!}{\color{blue}#1}{\!\!}}
\newcommand{\cellr}[1]{{\!\!}{\color{red}#1}{\!\!}}
\newcommand{\cell}[1]{{\!\!\!}#1{\!\!\!}}
\begin{center}
\begin{scriptsize}
\caption{A comparison between different methods for the non-composite setting \eqref{eq:std_prob} of \eqref{eq:reinforce_exp_prob}.}
\label{tbl:comp_compare}
\resizebox{0.49\textwidth}{!}{
\begin{tabular}{l ccc}\toprule
\cell{~~~~~~~~~ Algorithms} & \cell{Complexity} & \cell{Composite} & \cell{Single-loop}  \\ \midrule  
\cell{ REINFORCE \citep{Williams1992} } & \cell{$\BigO{\varepsilon^{-4}}$} & \cellr{\wontfix} &   \cellb{\done}  \\ \midrule
\cell{ GPOMDP \citep{Baxter2001} } & \cell{$\BigO{\varepsilon^{-4}}$} & \cellr{\wontfix} &   \cellb{\done} \\ \midrule
\cell{ SVRPG \citep{Papini2018a} } & \cell{$\BigO{\varepsilon^{-4}}$} & \cellr{\wontfix}  &  \cellr{\wontfix}  \\
\cell{ SVRPG \citep{Xu2019} } & \cell{$\BigO{\varepsilon^{-10/3}}$} & \cellr{\wontfix} &  \cellr{\wontfix} \\ \midrule
\cell{ HAPG \citep{Shen2019d} } & \cell{$\BigO{\varepsilon^{-3}}$} & \cellr{\wontfix} &  \cellr{\wontfix} \\ \midrule
\cell{ VRMPO \citep{Yang2019} } & \cell{$\BigO{\varepsilon^{-3}}$} & \cellr{\wontfix} &  \cellr{\wontfix} \\ \midrule
\cell{ SRVR-PG \citep{Xu2019SampleEP} } & \cell{$\BigO{\varepsilon^{-3}}$} & \cellr{\wontfix} &  \cellr{\wontfix} \\ \midrule
\cellb{\bf  This work} & \cellb{$\BigO{\varepsilon^{-3}}$} & \cellb{\done} &  \cellb{\done} \\ 
\bottomrule
\end{tabular}
}
\end{scriptsize}
\vspace{-2ex}
\end{center}
\end{table}

\paragraph{Our approach:}
Our approach lies in the stochastic variance reduction avenue, but using a completely new \textbf{hybrid} approach, leading to a novel estimator compared to existing methods in reinforcement learning. 
We build our estimator by taking a convex combination of the adapted SARAH \citep{Nguyen2017b} and REINFORCE \citep{Williams1992}, a classical unbiased policy gradient estimator.
This hybrid estimator not only allows us to trade-off the bias and variance between these two estimators but also possesses useful properties for developing new algorithms.
Note that the idea of combining stochastic estimators was first proposed for stochastic optimization in our recent works \citep{tran2019hybrid1,tran2019hybrid2}.
Unlike existing policy gradient methods, our algorithm first samples a large batch of trajectories to establish a good search direction.
After that, it iteratively updates the policy parameters using our hybrid estimator leading to a single-loop method without any snapshot loop as in SVRG or SARAH variants.
In addition, as regularization techniques have shown their effectiveness in deep learning \citep{Neyshabur2017,Zhang2017}, they possibly have great potential in reinforcement learning algorithms too.
A recent study \citep{liu2019regularization} shows that regularizations on the policy parameters can greatly improve the performance of policy gradient algorithms.
Motivated by these facts, we directly consider a new composite setting \eqref{eq:reinforce_exp_prob} as presented in Section~\ref{sec:main_theory}.
For this new composite model, it is not clear if existing algorithms remain convergent by simply adding a projection step on the constraint set, while our method does guarantee convergence.

\beforepara
\paragraph{Our contribution:}
To this end, our contribution in this paper can be summarized as follows:
\begin{compactitem}
\item[(a)] We introduce a novel hybrid stochastic policy gradient estimator by combining existing REINFORCE estimator with the adapted SARAH estimator for policy gradient.
We investigate some key properties of our estimator that can be used for algorithmic development.

\item[(b)] We propose a new algorithm to solve a composite maximization problem for policy optimization in reinforcement learning. 
Our model not only covers existing settings but also handles constraints and convex regularizers on policy parameters.

\item[(c)] We provide convergence analysis as the first theoretical result for composite optimization in reinforcement learning and estimate the trajectory complexity of our algorithm and show that our algorithm can achieve the best-known complexity over existing first-order methods (see Table~\ref{tbl:comp_compare}).

\end{compactitem}
Our algorithm only has one loop as REINFORCE or GPOMDP, which is fundamentally different from SVRPG, SVRG-adapted, and other SARAH-based algorithms for RL.
It can work with single sample or mini-batch and has two steps: proximal gradient step and averaging step with different step-sizes.
This makes the algorithm more flexible to use different step-sizes without sacrificing the overall complexity.

\beforepara
\paragraph{Paper outline:}
The rest of this paper is organized as follows.
Section~\ref{sec:prob_stat} describes problem of interest and gives an overview about policy gradient methods.
Section~\ref{sec:main_theory} introduces our new hybrid estimator for policy gradient and develops the main algorithm.
The complexity analysis is presented in Section~\ref{sec:convergence_analysis}, while Section~\ref{sec:experiments} provides several numerical examples.
All technical proofs and experimental details are given in Supplementary Document (Supp. Doc.).

\section{Model and Problem Statement}\label{sec:prob_stat}
\vspace{1ex}
\beforepara
\paragraph{Model:}
We consider a Markov Decision Process (MDP) \citep{Sutton2018} equipped with $6$ components $\{\Sc,\Ac,\Pc,\Rc, \gamma , \Pc_0\}$ where $\Sc$, $\Ac$ are the state and action spaces, $\Pc$ denotes the set of transition probabilities when taking certain actions, $\Rc$ is the reward function which characterizes the immediate reward earned by taking certain action, $\gamma$ is a discount factor, and $\Pc_0$ is the initial state distribution. 

Let $\pi( \cdot \vert s)$ be a density function over $\Ac$ when current state is $s$ and $\pi_\theta(\cdot \vert s)$ is a policy parameterized by parameter $\theta$.
A trajectory $\tau =\{s_0,a_0,s_1,a_1,\cdots,s_{H-1},a_{H-1} \}$ with effective length $H$ is a collection of states and actions sampled from a stationary policy. 
Denote $p_\theta(\cdot)$ as the density induced by policy $\pi_\theta$ over all possible trajectories and $p_\theta(\tau)$ is the probability of observing a trajectory $\tau$.
Also, let $\Rc(\tau) = \sum_{t=0}^{H-1} \gamma^t \Rc(s_t,a_t)$ be the total discounted reward for a trajectory $\tau$. 
Solving an MDP is equivalent to finding the solution that maximizes the expected cumulative discounted rewards.

\beforepara
\paragraph{Classical policy gradient methods:}
Policy gradient methods seek a differentiable parameterized policy $\pi_\theta$ that maximizes the expected cumulative discounted rewards as
\begin{equation}\label{eq:std_prob}
\max_{\theta \in \R^q} \Big\{ J(\theta) := \Exps{\tau \sim p_\theta}{\Rc(\tau)} \Big\}.
\end{equation}
where $q$ is the parameter dimension. The policy gradient theorem \citep{Sutton1999} shows that
\begin{equation*}
\nabla J(\theta) = \Exps{\tau \sim p_\theta}{ \nabla \log p_\theta(\tau) \Rc(\tau)},
\end{equation*}
where the policy gradient does not depend on the gradient of the state distribution despite the fact that the state distribution depends on the policy parameters \citep{Silver2014}.

This policy gradient can be used in gradient ascent algorithms to update the parameter $\theta$. 
However, we cannot calculate the full gradient at each update as we only get a finite number of samples at each iteration. Consequently, the policy gradient is often estimated by its sample average.
At each iteration, a batch of trajectories $\Bc = \{\tau_i\}_{i=1,\cdots,N}$ will be sampled from the environment to estimate the policy gradient as
\begin{equation*}
\widetilde{\nabla} J(\theta) := \dfrac{1}{N} \sum_{i=1}^N g(\tau_i\vert\theta),
\end{equation*}
where $g(\tau_i\vert\theta)$ is a sample estimator of $\Exps{\tau_i \sim p_\theta}{\nabla \log p_\theta(\tau_i) \Rc(\tau_i)}$.
We call $\widetilde{\nabla}{J}(\theta)$ a stochastic policy gradient (SPG) estimator. 
This estimator has been exploited in the two well-known REINFORCE \citep{Williams1992} and GPOMDP \citep{Baxter2001} methods.
The main step of policy gradient ascent methods is to update the parameters as
\begin{equation*}
\theta_{t+1} := \theta_t + \eta \nabla J(\theta_t), ~t = 0, 1, \cdots,
\end{equation*}
where $\eta > 0$ is some appropriate learning rate, which can be fixed or varied over $t$.
Since the policy changes after each update, the density $p_\theta(\cdot)$ also changes and creates non-stationarity in the problem which will be handled by importance weight in Section \ref{sec:main_theory}.

\section{A New Hybrid Stochastic Policy Gradient Algorithm}\label{sec:main_theory}
In this section, we first introduce a composite model for policy optimization.
Next, we extend the hybrid gradient idea from \cite{tran2019hybrid2} to policy gradient estimators.
Finally, we develop a new proximal policy gradient algorithm and its restart variant to solve the composite policy optimization problem and analyze their trajectory complexity.

\subsection{Composite Policy Optimization Model}
While the objective function in \eqref{eq:std_prob} is standard in most policy gradient methods, it is natural to have some constraints or regularizers on the policy parameters. 
In addition, adding constraints can prevent the explosion of parameters in highly nonlinear models as often seen in deep learning \citep{srivastava2014dropout}.
Adopting the idea of composite nonconvex optimization \citep{Pham2019ProxSARAH}, we are interested in the more general optimization problem in reinforcement learning as follow:
\begin{equation}\label{eq:reinforce_exp_prob}
{\!\!\!\!}\max_{\theta \in \R^q} \Big\{J(\theta) - Q(\theta) =  \Exps{\tau \sim p_\theta}{\Rc(\tau)} - Q(\theta) \Big\},{\!\!\!\!}
\end{equation}
where $Q(\theta)$ is a proper, closed, and convex function acting as a regularizer which can be the indicator function of a convex set representing the constraints on the parameters or some standard regularizers such as $\ell_1$-norm or  $\ell_2$-norm.
If there is no regularizer $Q(\theta)$, the problem \eqref{eq:reinforce_exp_prob} reduces to the standard one in \eqref{eq:std_prob}.

\subsection{Assumptions}
Let $F(\theta) := J(\theta) - Q(\theta)$ be the total objective function. 
We impose the following assumptions for our convergence analysis, which are often used in practice.

\begin{assumption}\label{ass:A1}
The regularizer $Q: \R^q \to \R\cup \{ +\infty\}$ is a proper, closed, and convex function. We also assume that the domain of $F$ is nonempty and there exists a finite upper bound
\begin{equation*}
{\!\!\!\!}F^* := \sup_{\theta \in \R^q} \left\{F(\theta) := J(\theta) - Q(\theta) \right\} < +\infty.{\!\!\!\!}
\end{equation*}
\end{assumption}

\begin{assumption}\label{ass:A2}
The immediate reward function is bounded, i.e., there exists $R > 0$ such that for all $a \in \Ac$, $s\in \Sc$, 
$\abs{\Rc(s,a)} \le R.$
\end{assumption}

\begin{assumption}\label{ass:A3}
Let $\pi_\theta(s,a)$ be the policy for a given state-action pair $(s,a)$. 
Then, there exist two positive constants $G$ and $M$ such that
\begin{equation*}
\|\nabla \log \pi_\theta(s,a)\| \le G \text{ and }\|\nabla^2 \log \pi_{\theta}(s,a) \| \le M,
\end{equation*}
for any $a \in \Ac$, $s \in \Sc$ where $\norm{\cdot}$ is the $\ell_2$-norm.
\end{assumption}
This assumption leads to useful results about the smoothness of $J(\theta)$ and $g(\tau\vert\theta)$ and the upper bound on the variance of the policy gradient estimator.

\begin{lemma}[\citep{Papini2018a,Shen2019d,Xu2019}]\label{lem:smoothness}
Under Assumption~\ref{ass:A2} and \ref{ass:A3}, for all $\theta,\theta_1,\theta_2\in\R^q$, we have
\begin{compactitem}
\item $\Vert \nabla{J}(\theta_1) - \nabla{J}(\theta_2)\Vert \leq L\norms{\theta_1 - \theta_2}$;
\vspace{0.75ex}
\item $\norm{g(\tau\vert\theta_1) - g(\tau \vert\theta_2)} \le L_g \norm{\theta_1 - \theta_2}$;
\vspace{0.75ex}
\item $\norm{g(\tau,\theta)} \le C_g $; and
\vspace{0.75ex}
\item $\norm{g(\tau\vert\theta) - \nabla J(\theta)}^2 \le \sigma^2$,
\end{compactitem}
where $g(\cdot)$ is the REINFORCE estimator and $L$, $L_g$, $C_g$, and $\sigma^2$ are constants depending only on $R$, $G$, $M$, $H$, $\gamma$, and the baseline $b$.
\end{lemma}
For more details about the constants and the proofs of Lemma~\ref{lem:smoothness} we refer e.g., to \cite{Papini2018a,Shen2019d,Xu2019}.

\begin{assumption}\label{ass:A4}
There exists a constant $W \in (0,\infty)$ such that, for each pair of policies encountered in Algorithm~\ref{alg:A1} the following holds
\begin{equation*}
{\!\!\!\!}\Var{\omega(\tau\vert\theta_1,\theta_2)} \le W,~~~\theta_1,\theta_2 \in \R^q,~\tau \sim p_{\theta_1},{\!\!\!\!}
\end{equation*}
where $\omega(\tau\vert\theta_1,\theta_2)=\frac{p_{\theta_2}(\tau)}{p_{\theta_1}(\tau)}$ is the importance weight between $p_{\theta_2}(\cdot)$ and $p_{\theta_1}(\cdot)$.
\end{assumption}

Since the importance weight  $\omega$ introduces another source of variance, we require this assumption for our convergence analysis as used in previous works, e.g., in \cite{Papini2018a,Xu2019}.

\begin{remark}\label{rem:gaussian}
\cite{Cortes2010} shows that if $\sigma_Q$, $\sigma_P$ are variances of two Gaussian distributions P and Q, and $\sigma_Q > \frac{1}{\sqrt{2}}\sigma_P$ then the variance of the importance weights is bounded, i.e. Assumption \ref{ass:A4} holds for Gaussian policies which are commonly used to represent the policy in continuous control tasks.
\end{remark}

\subsection{Optimality Condition}
Associated with problem \eqref{eq:reinforce_exp_prob}, we define
\begin{equation}\label{eq:grad_map_def}
{\!\!\!\!}\Gc_{\eta}(\theta) := \eta^{-1}\left[ \prox_{\eta Q}\left( \theta + \eta \nabla J(\theta)) - \theta \right)\right],
\end{equation}
for some $\eta > 0$ as the gradient mapping of $F(\theta)$  \citep{Nesterov2014}, where $\prox_{Q}(\theta) := \argmin_{\theta'} \{Q(\theta') + \tfrac{1}{2}\norm{\theta' - \theta}^2 \}$ denotes the proximal operator of $Q$ (see, e.g.,  \cite{Parikh2014} for more details).

A point $\theta^{*}$ is called a stationary point of \eqref{eq:reinforce_exp_prob} if 
\begin{equation*}\label{eq:opt_cond}
\Exp{\norms{\Gc_{\eta}(\theta^{\ast})}^2} = 0.
\end{equation*}

Our goal is to design an iterative method to produce an $\varepsilon$-approximate stationary point $\tilde{\theta}_T$ of \eqref{eq:reinforce_exp_prob} after at most $T$ iterations defined as
\begin{equation*}
\Exp{\norms{\Gc_{\eta}(\tilde{\theta}_T)}^2} \le \varepsilon^2,
\end{equation*}
where $\varepsilon > 0$ is a desired tolerance, and the expectation is taken overall the randomness up to $T$ iterations.

\subsection{Novel Hybrid SPG Estimator}
\paragraph{Unbiased estimator:}
Recall that given a trajectory $\tau := \set{s_0,a_0,\cdots, s_{H-1},a_{H-1} }$, the REINFORCE (SPG) estimator is defined as
\begin{equation*}
\begin{array}{lll}
{\!\!\!\!\!}g(\tau\vert\theta) &:= \left[\dsum_{t=0}^{H-1} \nabla \log \pi_\theta(a_t \vert s_t) \right] \Rc(\tau), {\!\!\!\!\!}
\end{array}
\end{equation*}
where $\Rc(\tau) :=  \sum_{t=0}^{H-1}\gamma^{t} \Rc(s_t,a_t)$.

Note that the REINFORCE estimator is unbiased, i.e. $\Exps{\tau \sim p_\theta}{g(\tau\vert\theta)} = \nabla J(\theta)$. 
In order to reduce the variance of these estimators, a baseline is normally added while maintaining the unbiasedness of the estimators \citep{Sutton2018,Zhao2011}. From now on, we will refer to $g(\tau\vert\theta)$ as the baseline-added version defined as
\begin{equation*}
\begin{array}{ll}
g(\tau\vert\theta) &:= \sum_{t=0}^{T-1} \nabla \log \pi_\theta(a_t \vert s_t) A_t  ,
\end{array}
\end{equation*}
where $A_t := \Rc(\tau) - b_t$ with $b_t$ being a baseline and possibly depending only on $s_t$.

\paragraph{Hybrid SPG estimator:}
In order to reduce the number of trajectories sampled, we extend our idea in \cite{tran2019hybrid2} for stochastic optimization to develop a new hybrid stochastic policy gradient (HSPG) estimator that helps balance the bias-variance trade-off. 
The estimator is formed by taking a convex combination of two other estimators: one is an unbiased estimator which can be REINFORCE estimator, and another is the adapted SARAH estimator \citep{Nguyen2017b} for policy gradient which is biased.

More precisely, if $\Bc_t$ and $\widehat{\Bc}_t$ are two random batches of trajectories with sizes $B$ and $\widehat{B}$, respectively, sampled from $p_{\theta_t}(\cdot)$, the hybrid stochastic policy gradient estimator at $t$-th iteration can be expressed as
\begin{equation}\label{eq:hybrid_est}
\begin{array}{ll}
v_t &:= \beta v_{t-1} + \frac{\beta}{B}\displaystyle\sum_{\tau \in \Bc_t} \Delta g(\tau\vert \theta_t) \\
&{~~~} + {~} \frac{(1-\beta) }{\widehat{B}}\displaystyle\sum_{\hat{\tau} \in \widehat{\Bc}_t} g(\hat{\tau} \vert \theta_t),
\end{array}
\end{equation}
where 
\begin{equation*}
\Delta{g}(\tau\vert \theta_t) :=  g(\tau\vert \theta_t) - \omega(\tau \vert \theta_t,\theta_{t-1}) g(\tau \vert \theta_{t-1}),
\end{equation*}
and
\begin{equation*}
v_0 := \dfrac{1}{N}\dsum_{\tau\in\tilde{\Bc}}g(\tau|\theta_0),
\end{equation*}
with $\tilde{\Bc}$ is a batch of trajectories collected at the beginning.
Note that $\omega(\tau\vert\theta_t,\theta_{t-1})$ is an importance weight added to account for the distribution shift since the trajectories $\tau \in \Bc_t$ are sampled from $p_{\theta_t}(\cdot)$ but not from $p_{\theta_{t-1}}(\cdot)$. 
Note also that $v_t$ in \eqref{eq:hybrid_est} is also different from the momentum SARAH estimator recently proposed in \cite{cutkosky2019momentum}.

\subsection{The Complete Algorithm}
The novel Proximal Hybrid Stochastic Policy Gradient Algorithm (abbreviated by ProxHSPGA) to solve \eqref{eq:reinforce_exp_prob} is presented in detail in Algorithm~\ref{alg:A1}.

\begin{algorithm}[ht!]\caption{(ProxHSPGA)}\label{alg:A1}
\small
\begin{algorithmic}[1]
   \State{\bfseries Initialization:} An initial point $\theta_0\in\R^q$, and positive parameters $m$, $N$, $B$, $\widehat{B}$, $\beta$, $\alpha$, and $\eta$ (specified later).
   \vspace{0.5ex}
   \State\hspace{0ex} Sample a batch of trajectories $\tilde{\Bc}$ of size $N$ from $p_{\theta_0}(\cdot)$.
   \State\hspace{0ex} Calculate $v_0 := \dfrac{1}{N}\dsum_{\tau\in\tilde{\Bc}}g(\tau|\theta_0)$.
  \State\hspace{0ex}\label{step:o4} Update 
  \vspace{-1.5ex}
  \begin{equation*}
  \left\{\begin{array}{ll}
  \widehat{\theta}_1 &:= \prox_{\eta Q}(\theta_0 + \eta v_0) \vspace{0.5ex}\\
  \theta_1 &:= (1-\alpha)\theta_0 + \alpha \widehat{\theta}_1.
   \end{array}\right.
  \end{equation*}
  \vspace{-2.5ex}
   \State\hspace{0ex}{\bfseries For $t := 1,\cdots,m$ do}
   \vspace{0.5ex}   
   \State\hspace{2ex} Generate $2$ independent batches of trajectories $\Bc_t$ and $\widehat{\Bc}_t$ with size $B$ and $\hat{B}$ from $p_{\theta_t}(\cdot)$.{\!\!\!}
   \State\hspace{2ex} Evaluate the hybrid estimator $v_t$ as in \eqref{eq:hybrid_est}.
   \vspace{0.5ex}   
   \State\hspace{2ex} Update
   \vspace{-1.5ex}
  \begin{equation*}
  \left\{\begin{array}{ll}
  \widehat{\theta}_{t+1} &:= \prox_{\eta Q}(\theta_t + \eta v_t) \vspace{0.5ex}\\
  \theta_{t+1} &:= (1-\alpha)\theta_t + \alpha \widehat{\theta}_{t+1}.
   \end{array}\right.
  \end{equation*}
  \vspace{-2.5ex}
   \State\hspace{0ex}{\bfseries EndFor}
   \vspace{0.5ex}   
   \State\hspace{0ex}Choose $\widetilde{\theta}_T$ from $\left\{\theta_t\right\}_{t=1}^m$ uniformly randomly. 
\end{algorithmic}
\end{algorithm}

Unlike SVRPG \citep{Papini2018a,Xu2019} and HAPG \citep{Shen2019d}, Algorithm~\ref{alg:A1} only has \textbf{one loop} as REINFORCE or GPOMDP. 
Moreover, Algorithm~\ref{alg:A1} does not use the estimator for the policy Hessian as in HAPG.
At the initial stage, a batch of trajectories is sampled using $p_{\theta_{0}}$ to estimate an initial policy gradient estimator which provides a good initial search direction. 
At the $t$-th iteration, two independent batches of trajectories are sampled from $p_{\theta_t}$ to evaluate the hybrid stochastic policy gradient estimator. 
After that, a proximal step followed by an averaging step are performed which are inspired by \cite{Pham2019ProxSARAH}.
Note that the batches of trajectories at each iteration are sampled from the current distribution which will change after each update.
Therefore, the importance weight $\omega(\tau\vert\theta_t,\theta_{t-1})$ is introduced to account for the non-stationarity of the sampling distribution.
As a result, we still have $\Exps{\tau\sim p_{\theta_t}}{\omega(\tau\vert\theta_t,\theta_{t-1})g(\tau \vert \theta_{t-1})} = \nabla J(\theta_{t-1})$.

\subsection{Restarting variant}
While Algorithm~\ref{alg:A1} has the best-known theoretical complexity as shown in Section~\ref{sec:convergence_analysis}, its practical performance may be affected by the constant step-size $\alpha$ depending on $m$. 
As will be shown later, the step-size $\alpha \in [0,1]$ is inversely proportional to the number of iterations $m$ and it is natural to have $\alpha$ close to $1$ to take advantage of the newly computed information. 
To increase the practical performance of our algorithm without sacrificing its complexity, we propose to inject a simple restarting strategy by repeatedly running Algorithm~\ref{alg:A1} for multiple stages as in Algorithm~\ref{alg:A2}.

\begin{algorithm}[ht!]\caption{(Restarting ProxHSPGA)}\label{alg:A2}
\small
\begin{algorithmic}[1]
\vspace{-0.5ex}
   \State{\bfseries Initialization:} Input an initial point $\theta^{(0)}_0$.
   \State\hspace{0ex}{\bfseries For $s := 0,\cdots,S-1$ do}
   \State\hspace{3ex} Run Algorithm~\ref{alg:A1} with $\theta_0 := \theta_0^{(s)}$.
   \State\hspace{3ex} Output $\theta_0^{(s+1)} := \theta_{m+1}$.
   \State\hspace{0ex}{\bfseries EndFor}
   \State\hspace{0ex}Choose $\widetilde{\theta}_T$ uniformly randomly from $\{\theta_t^{(s)}\}_{t=0\to m}^{s=0\to S-1}$. {\!\!\!\!}
\end{algorithmic}
\end{algorithm}

We emphasize that without this restarting strategy, Algorithm~\ref{alg:A1} still converges and the restarting loop in Algorithm~\ref{alg:A2} does not sacrifice the best-known complexity as stated in the next section.

\section{Convergence Analysis}\label{sec:convergence_analysis}
This section presents key properties of the hybrid stochastic policy gradient estimators as well as the theoretical convergence analysis and complexity estimate.

\subsection{Properties of the hybrid SPG estimator}
Let $\Fc_t := \sigma\left(\tilde{\Bc},\Bc_1, \widehat{\Bc}_1,\cdots,\Bc_{t-1}, \widehat{\Bc}_{t-1}\right)$ be the $\sigma$-field generated by all trajectories sampled up to the $t$-th iteration. 
For the sake of simplicity, we assume that $B = \widehat{B}$ but our analysis can be easily extended for the case $B \neq \widehat{B}$. 
Then the hybrid SPG estimator $v_t$ has the following properties
\begin{lemma}[Key properties]\label{lem:prop_hybrid}
Let $v_t$ be defined as in \eqref{eq:hybrid_est} and $\Delta{v}_t :=  v_t - \nabla J(\theta_t)$. Then
\begin{equation}\label{eq:lem41_1}
\Exps{\tau,\hat{\tau} \sim p_{\theta_t}}{v_t} = \nabla J(\theta_t) + \beta \Delta{v}_{t-1}.
\end{equation}
If $\beta \neq 0$, then $v_t$ is an biased estimator. 
In addition, we have
\begin{equation}\label{eq:lem41_2}
{\!\!\!\!\!\!\!}\begin{array}{ll}
\Exps{\tau,\hat{\tau} \sim p_{\theta_t}}{\norms{\Delta v_t}^2}  {\!\!\!\!\!\!}&\le \beta^2 \|\Delta v_{t-1}\|^2 +\frac{(1-\beta)^2\sigma^2}{B}\\
&~~~~ + {~} \frac{\beta^2\overline{C}}{B}\norm{\theta_t - \theta_{t-1}}^2,
\end{array}{\!\!\!\!\!\!}
\end{equation}
where $\overline{C} > 0$ is a given constant.
\end{lemma}
The proof of Lemma~\ref{lem:prop_hybrid} and the explicit constants are given in  Supp. Doc.~\ref{app:A1} due to space limit.

\subsection{Complexity Estimates}

The following lemma presents a key estimate for our convergence results.
\begin{lemma}[One-iteration analysis]\label{lem:key_estimate}
Under Assumption~\ref{ass:A2}, \ref{ass:A3}, and \ref{ass:A4}, let $\{\widehat{\theta}_t,\theta_t\}_{t=0}^m$ be the sequence generated by Algorithm~\ref{alg:A1} and $\Gc_\eta$ be the gradient mapping defined in \eqref{eq:grad_map_def}.
Then
\begin{equation}\label{eq:lem43}
{\!\!\!\!\!\!}\begin{array}{ll}
&\Exp{F(\theta_{t+1})}  \ge \Exp{F(\theta_t) }+ \frac{\eta^2 \alpha}{2} \Exp{\norms{\Gc_\eta(\theta_t)}^2} \vspace{1ex}\\
&{~~~~~~} - {~} \frac{\xi}{2}\Exp{\norms{v_{t} \!-\! \nabla J(\theta_{t})}^2} \!+\! \frac{\zeta}{2}\Exp{\norms{\widehat{\theta}_{t\!+\!1} \!-\! \theta_t}^2},
\end{array}{\!\!\!\!\!\!}
\end{equation}
where $\xi := \alpha(1 + 2\eta^2)$ and $\zeta := \alpha\big(\frac{2}{\eta} - L\alpha - 3\big) > 0$ provided that $\alpha\in (0,1]$ and $\frac{2}{\eta} - L\alpha - 3 > 0$.
\end{lemma}

The following theorem summarizes the convergence analysis of Algorithm~\ref{alg:A1}.
\begin{theorem}\label{thm:grad_bound}
Under Assumption~\ref{ass:A1}, \ref{ass:A2}, \ref{ass:A3}, and \ref{ass:A4}, let $\left\{\theta_{t}\right\}_{t=0}^{m}$ be the sequence generated by Algorithm~\ref{alg:A1} with
\begin{equation}\label{eq:step_sizes}
\left\{\begin{array}{ll}
\beta := 1 - \frac{\sqrt{B}}{\sqrt{N(m+1)}} \\
\alpha := \frac{\hat{c} \sqrt{2}B^{3/4}}{\sqrt{3\overline{C}} N^{1/4}(m+1)^{1/4}}\\
\eta := \frac{2}{4 + L\alpha},
\end{array}\right.,
\end{equation}
where $B,\hat{c},L,$ and $\overline{C}$ are given constants.
If $\tilde{\theta}_T$ is chosen uniformly at random from $\left\{\theta_{t}\right\}_{t=0}^{m}$, then the following estimate holds
\begin{equation}\label{eq:thm_grad_bound}
{\!\!\!\!}\begin{array}{ll}
& \Exp{\norms{\Gc_\eta(\tilde{\theta}_T)}^2} \le \dfrac{3(4 + L)^2 \sigma^2}{ 4[BN(m+1)]^{1/2}} \vspace{1ex}\\
&{~~~~~~~~~~~~} + {~} \dfrac{(4 + L)^2\sqrt{3\overline{C}}N^{1/4}}{4 \hat{c}\sqrt{2}[B(m+1)]^{3/4}}\left[ F^* - F(\theta_0) \right].
\end{array}{\!\!\!\!}
\end{equation}
\end{theorem}

Consequently, the trajectory complexity is presented in the following corollary.
\begin{corollary}\label{cor:complexity}
For both Algorithm~\ref{alg:A1} and Algorithm~\ref{alg:A2}, 
let us fix $B \in \mathbb{N}_+$ and set $N := \tilde{c}\sigma^{8/3}[B(m+1)]^{1/3}$ for some $\tilde{c} > 0$ in Theorem~\ref{thm:grad_bound}. If we also choose $m$ in Algorithm~\ref{alg:A1} such that $m+1 = \frac{\Psi_0^{3/2}\sigma}{B\varepsilon^3}$ and choose $m,S$ in Algorithm~\ref{alg:A2} such that $S(m+1) = \frac{\Psi_0^{3/2}\sigma}{B\varepsilon^3}$ for some constant $\Psi_0$, then the number of trajectories $\Tc_{\text{traj}}$ to achieve $\tilde{\theta}_T$ such that $\Exp{\norms{\Gc_\eta(\tilde{\theta}_T)}^2} \le \varepsilon^2$ for any $\varepsilon > 0$ is at most
\begin{equation*}
\Tc_{\text{traj}} = \BigO{\varepsilon^{-3}}.
\end{equation*}
where $\tilde{\theta}_T$ is chosen uniformly at random from $\{\theta^{(s)}_{t}\}_{t=0,\cdots,m}^{s=0,\cdots,S-1}$ if using Algorithm~\ref{alg:A2}.
\end{corollary}

The proof of Theorem~\ref{thm:grad_bound} and Corollary~\ref{cor:complexity} are given in Supp. Doc.~\ref{app:A3}, and \ref{app:A4}, respectively.

Comparing our complexity bound with other existing methods in Table~\ref{tbl:comp_compare}, we can see that we improve a factor of $\boldsymbol{\varepsilon^{-1/3}}$ over SVRPG in \cite{Xu2019} while matching the best-known complexity without the need of using the policy Hessian estimator as HAPG from \cite{Shen2019d}.

\section{Numerical Experiments}\label{sec:experiments}
In this section, we present three examples to provide comparison between the performance of HSPGA and other related policy gradient methods.
We also provide an example to illustrate the effect of the regularizer $Q(\cdot)$ to our model \eqref{eq:reinforce_exp_prob}.
More examples can be found in the Supp. Doc. \ref{app:num_exp}.
All experiments are run on a Macbook Pro with 2.3 GHz Quad-Core, 8GB RAM.

\begin{figure}[htp!]
\vspace{-2ex}
\begin{center}
\includegraphics[width = 0.485\textwidth]{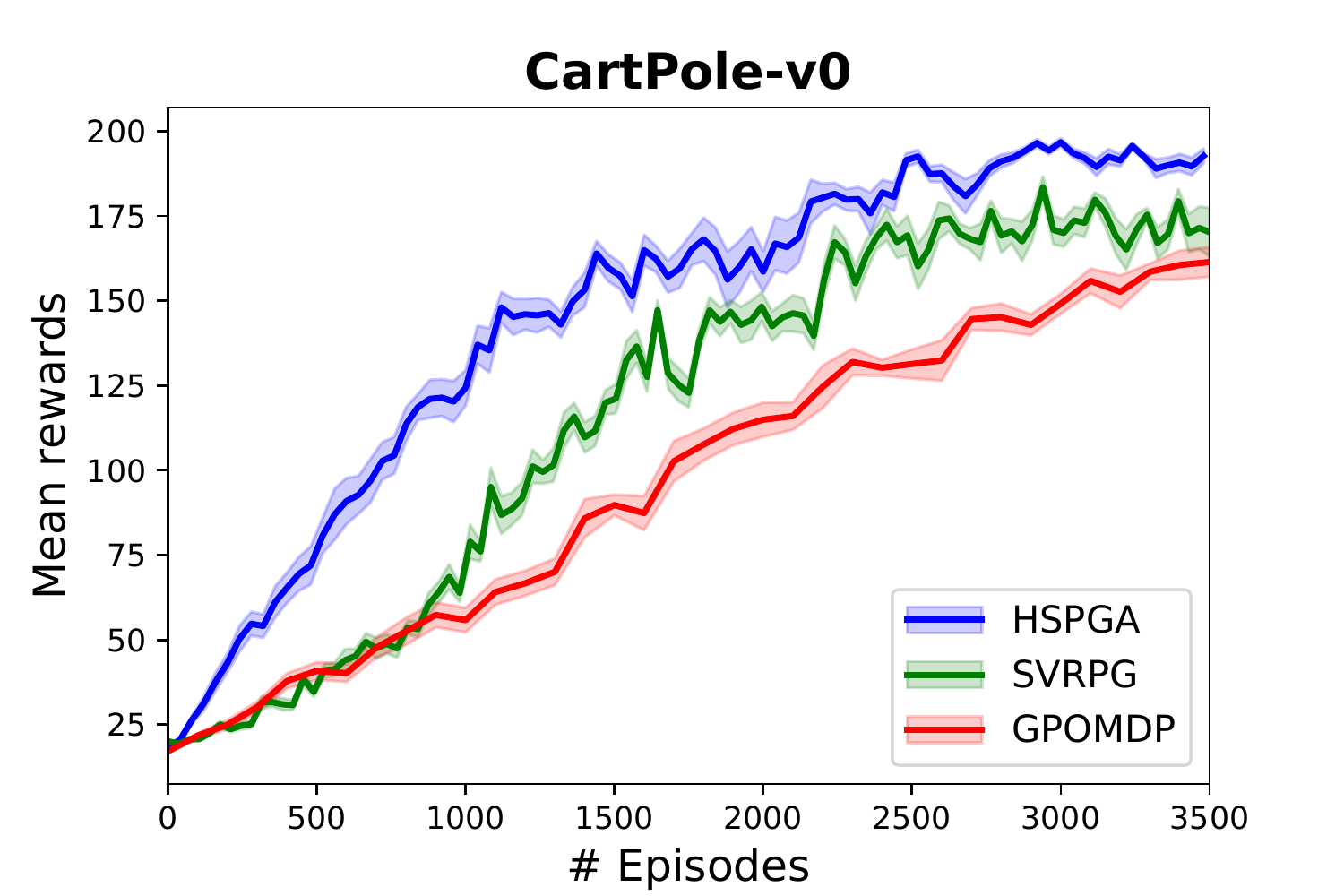}
\vspace{-4ex}
\caption{The performance of three algorithms on the \texttt{Carpole-v0} environment.}\label{fig:cartpole}
\end{center}
\vspace{-2ex}
\end{figure}
\begin{figure}[htp!]
\begin{center}
\includegraphics[width = 0.485\textwidth]{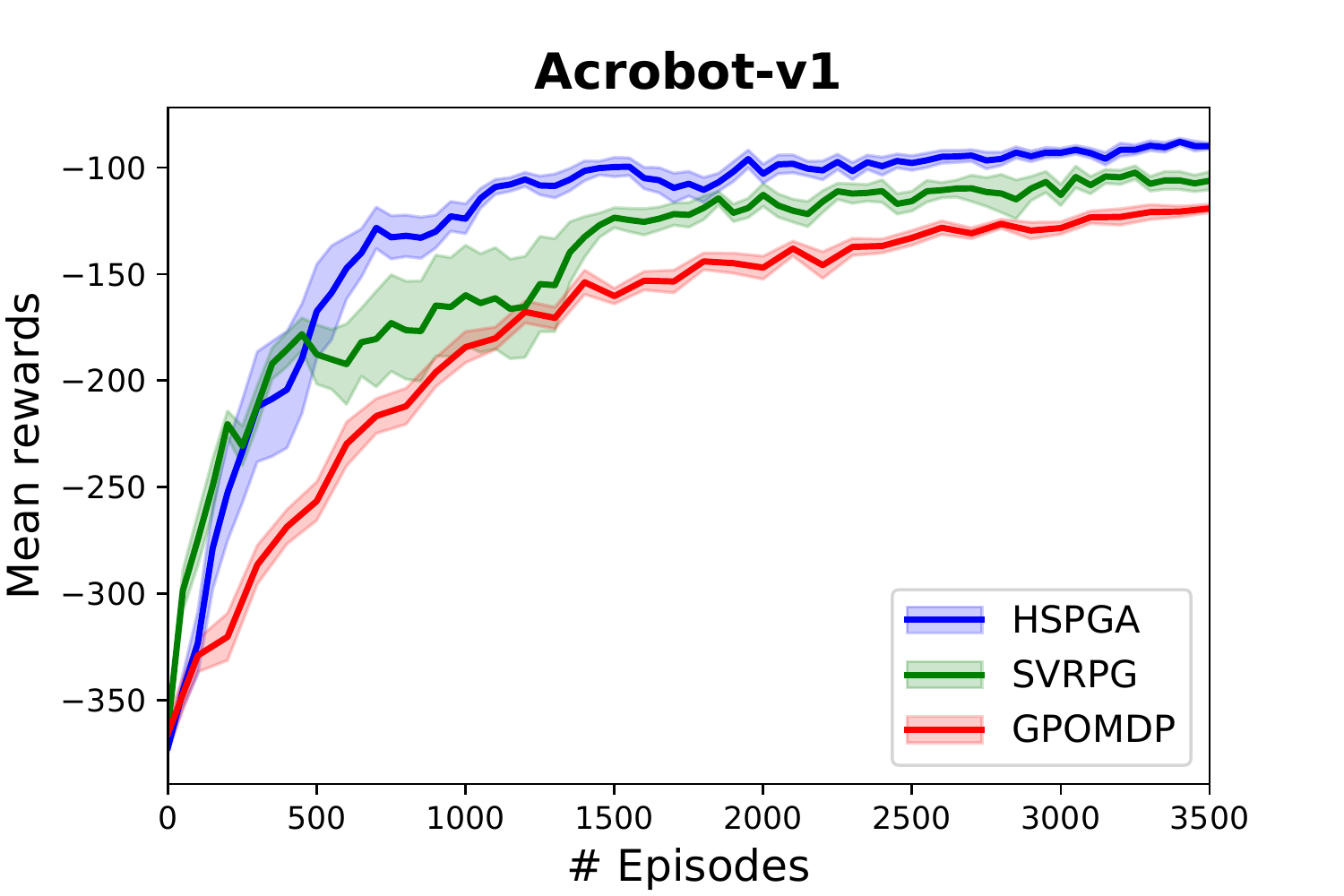}
\vspace{-4ex}
\caption{The performance of three algorithms on the \texttt{Acrobot-v1} environment.}\label{fig:acrobot}
\vspace{-3ex}
\end{center}
\end{figure}

We implement our restarting algorithm, Algorithm \ref{alg:A2}, on top of the \texttt{rllab}\footnote{Available at \href{https://github.com/rll/rllab}{https://github.com/rll/rllab}} library \citep{Duan2016}. 
The source code is available at \href{https://github.com/unc-optimization/ProxHSPGA}{https://github.com/unc-optimization/ProxHSPGA}.
We  compare our algorithm with two other methods: SVPRG \citep{Papini2018a, Xu2019} and GPOMDP \citep{Baxter2001}. 
Although REINFORCE and GPOMDP have the same trajectory complexity, as observed in \citep{Papini2018a}, GPOMDP often performs better than REINFORCE, so we only choose to implement GPOMDP in our experiments. 
Since SVRPG and GPOMDP solves the non-composite problems \eqref{eq:std_prob}, we set $Q(\theta) = 0$ in the first three examples and adjust our algorithm, denoted as HSPGA, accordingly. We compare our algorithm with the fixed epoch length variant of SVRPG as reported in \cite{Papini2018a, Xu2019}. 
For the implementation of SVRPG and GPOMDP, we reuse the implementation of Papini et al.\footnote{Available at \href{https://github.com/Dam930/rllab}{https://github.com/Dam930/rllab}}. 
We test these algorithms on three well-studied reinforcement learning tasks: \texttt{Cart Pole}, \texttt{Acrobot}, and \texttt{Moutain Car} which are available in OpenAI \texttt{gym} \citep{OpenAIGym}, a well-known toolkit for developing and comparing reinforcement learning algorithms.  We also test these algorithms on continuous control tasks using other simulators such as \texttt{Roboschool} \citep{roboschool} and \texttt{Mujoco} \citep{Todorov2012MuJoCoAP}.

For each environment, we initialize the policy randomly and use it as initial policies for all $10$ runs of all algorithms. 
The performance measure, i.e., mean rewards, is computed by averaging the final rewards of $50$ trajectories sampled by the current policy. 
We then compute the mean and 90\% confidence interval across $10$ runs of these performance measures at different time point. 
In all plots, the solid lines represent the mean and the shaded areas are the confidence band of the mean rewards. 
In addition, detailed configurations of the policy network and parameters can be found in Supp. Doc.~\ref{sec:exp_conf}. 
We note that the architecture of the neural network is denoted as $[\text{observation space}] \times [\text{hidden layers}] \times [\text{action space}]$. 

\beforepara
\paragraph{Cart Pole-v0 environment:}
For the \texttt{Cart pole} environment, we use a deep soft-max policy network \citep{Bridle1990, BerkeleyRL,Sutton2018} with one hidden layer of $8$ neurons. 
Figure~\ref{fig:cartpole} depicts the results where we run each algorithm for $10$ times and compute the mean and $90\%$ confidence intervals.

From Figure~\ref{fig:cartpole}, we can see that HSPGA outperforms the other $2$ algorithms while SVRPG works better than GPOMDP as expected. 
HSPGA is able to reach the maximum reward of $200$ in less than $4000$ episodes.

\begin{figure}[htp!]
\begin{center}
\vspace{-3ex}
\includegraphics[width = 0.485\textwidth]{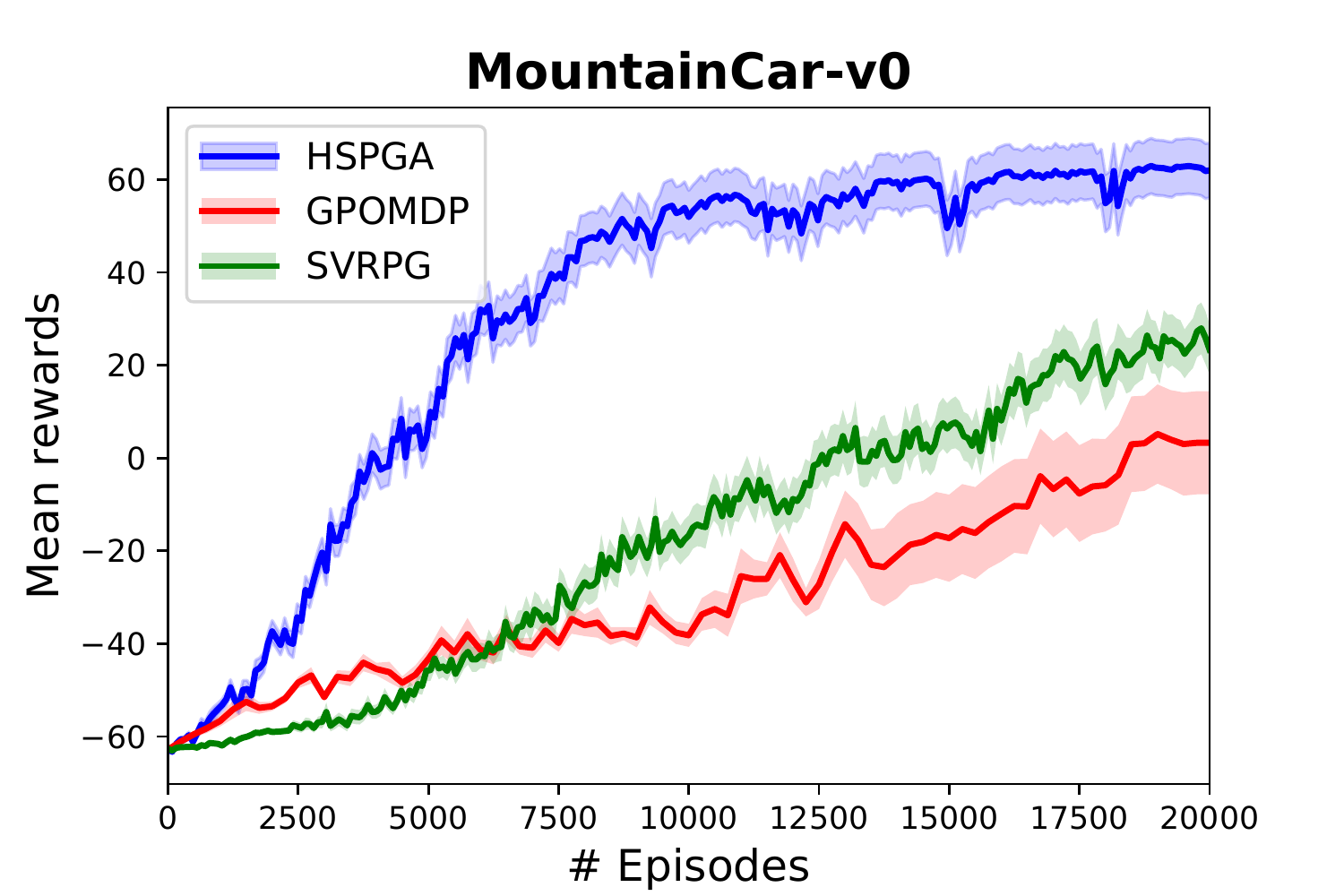}
\vspace{-4ex}
\caption{The performance of three algorithms on the \texttt{Mountain Car-v0} environment.}\label{fig:mountaincar}
\vspace{-1ex}
\end{center}
\end{figure}

\beforepara
\paragraph{Acrobot environment:}
Next, we evaluate three algorithms on the \texttt{Acrobot-v1} environment. 
Here, we use a deep soft-max policy with one hidden layer of 16 neurons. 
The performance of these 3 algorithms are illustrated in Figure~\ref{fig:acrobot}.

We observe similar results as in the previous example where HSPGA has the best performance over three candidates.
SVRPG is still better than GPOMDP in this example.

\beforepara
\paragraph{Mountain Car environment:}
For the \texttt{MountainCar-v0} environment, we use a deep Gaussian policy \citep{Sutton2018} where the mean is the output of a neural network containing one hidden layer of $8$ neurons and the standard deviation is fixed at $1$. 
The results of three algorithms are presented in Figure~\ref{fig:mountaincar}.

Figure~\ref{fig:mountaincar} shows that HSPGA highly outperforms the other two algorithms. 
Again, SVRPG remains better than GPOMDP as expected.

\beforepara
\paragraph{The effect of regularizers:}
We test the effect of the regularizer $Q(\cdot)$ by adding a Tikhonov one as
\begin{equation*}\label{eq:prob_l2reg}
{\!\!\!\!}\max_{\theta \in \R^q} \left\{J(\theta) - \lambda \norm{\theta}_2^2 \right\}. {\!\!\!\!}
\end{equation*}
This model was intensively studied in \cite{liu2019regularization}.

\begin{figure}[htp!]
\begin{center}
\vspace{-3ex}
\includegraphics[width = 0.485\textwidth]{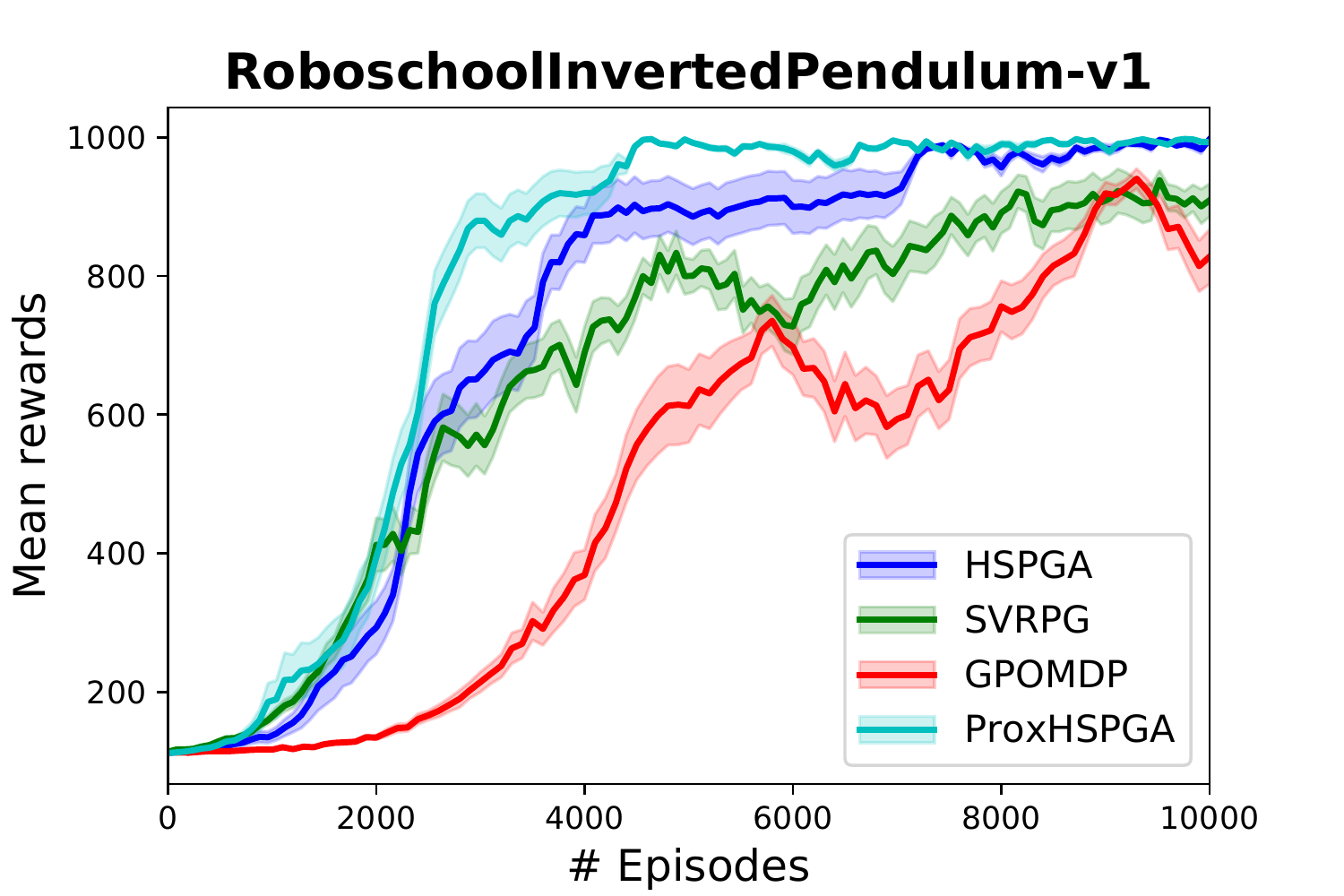}
\vspace{-4ex}
\caption{The performance of composite vs. non-composite algorithms on the \texttt{Roboschool Inverted Pendulum-v1} environment.}\label{fig:invpend}
\end{center}
\vspace{-1ex}
\end{figure}

We also compare all non-composite algorithms with ProxHSPGA in the \texttt{Roboschool Inverted Pendulum-v1} environment. 
In this experiment, we set the penalty parameter $\lambda = 0.001$ for ProxHSPGA.
The results are depicted in Figure~\ref{fig:invpend} and more information about the configuration of each algorithm is in Supp. Document~\ref{sec:exp_conf}.

From Figure~\ref{fig:invpend}, in terms of non-composite algorithms, HSPGA is the best followed by SVRPG and then by GPOMDP. 
Furthermore, ProxHSPGA shows its advantage by reaching the maximum reward of 1000 faster than HSPGA.

\beforesec
\section{Conclusion}
\aftersec
We have presented a novel policy gradient algorithm to solve regularized reinforcement learning models. 
Our algorithm uses a novel policy gradient estimator which is a combination of an unbiased estimator, i.e. REINFORCE estimator, and a biased estimator adapted from SARAH estimator for policy gradient. 
Theoretical results show that our algorithm achieves the best-known trajectory complexity to attain an $\varepsilon$-approximate first-order solution for the problem under standard assumptions. 
In addition, our numerical experiments not only help confirm the benefit of our algorithm compared to other closely related policy gradient methods but also verify the effectiveness of regularization in policy gradient methods.

\ackaccepted{
Q. Tran-Dinh has partly been supported by the National Science Foundation (NSF), grant no. DMS-1619884 and the Office of Naval Research (ONR), grant no. N00014-20-1-2088 (2020-2023).
Q. Tran-Dinh and N. H. Pham are partly supported by The Statistical and Applied Mathematical Sciences Institute (SAMSI).
}

\bibliographystyle{plainnat}
\bibliography{ref}

\begin{thebibliography}{62}
\providecommand{\natexlab}[1]{#1}
\providecommand{\url}[1]{\texttt{#1}}
\expandafter\ifx\csname urlstyle\endcsname\relax
  \providecommand{\doi}[1]{doi: #1}\else
  \providecommand{\doi}{doi: \begingroup \urlstyle{rm}\Url}\fi

\bibitem[Allen-Zhu(2017)]{Allen-Zhu2017}
Z.~Allen-Zhu.
\newblock Katyusha: The first direct acceleration of stochastic gradient
  methods.
\newblock In \emph{Proceedings of the 49th Annual ACM SIGACT Symposium on
  Theory of Computing}, pages 1200--1205, New York, NY, USA, 2017.

\bibitem[Allen-Zhu(2018)]{Allen-Zhu2018}
Z.~Allen-Zhu.
\newblock Natasha 2: Faster non-convex optimization than sgd.
\newblock In \emph{Advances in Neural Information Processing Systems 31}, pages
  2675--2686, 2018.

\bibitem[Allen-Zhu and Yuan(2016)]{Allen-Zhu2016}
Z.~Allen-Zhu and Y.~Yuan.
\newblock Improved svrg for non-strongly-convex or sum-of-non-convex
  objectives.
\newblock In \emph{Proceedings of the 33rd International Conference on
  International Conference on Machine Learning - Volume 48}, pages 1080--1089,
  2016.

\bibitem[Baxter and Bartlett(2001)]{Baxter2001}
J.~Baxter and P.~L. Bartlett.
\newblock Infinite-horizon policy-gradient estimation.
\newblock \emph{J. Artif. Int. Res.}, 15\penalty0 (1):\penalty0 319--350,
  November 2001.

\bibitem[Bridle(1990)]{Bridle1990}
J.~S. Bridle.
\newblock Training stochastic model recognition algorithms as networks can lead
  to maximum mutual information estimation of parameters.
\newblock In \emph{Advances in Neural Information Processing Systems 2}, pages
  211--217. Morgan-Kaufmann, 1990.

\bibitem[Brockman et~al.(2016)Brockman, Cheung, Pettersson, Schneider,
  Schulman, Tang, and Zaremba]{OpenAIGym}
G.~Brockman, V.~Cheung, L.~Pettersson, J.~Schneider, J.~Schulman, J.~Tang, and
  W.~Zaremba.
\newblock Openai gym, 2016.

\bibitem[Cortes et~al.(2010)Cortes, Mansour, and Mohri]{Cortes2010}
C.~Cortes, Y.~Mansour, and M.~Mohri.
\newblock Learning bounds for importance weighting.
\newblock In \emph{Advances in Neural Information Processing Systems 23}, pages
  442--450. 2010.

\bibitem[Cutkosky and Orabona(2019)]{cutkosky2019momentum}
A.~Cutkosky and F.~Orabona.
\newblock Momentum-based variance reduction in non-convex sgd.
\newblock In \emph{Advances in Neural Information Processing Systems}, pages
  15210--15219, 2019.

\bibitem[DeepMind(2019)]{alphastarblog}
DeepMind.
\newblock {AlphaStar: Mastering the Real-Time Strategy Game StarCraft II}.
\newblock
  \href{https://deepmind.com/blog/alphastar-mastering-real-time-strategy-game-starcraft-ii/}{https://deepmind.com/blog},
  2019.

\bibitem[Defazio et~al.(2014)Defazio, Bach, and Lacoste-Julien]{Defazio2014}
A.~Defazio, F.~Bach, and S.~Lacoste-Julien.
\newblock Saga: A fast incremental gradient method with support for
  non-strongly convex composite objectives.
\newblock In \emph{Proceedings of the 27th International Conference on Neural
  Information Processing Systems - Volume 1}, pages 1646--1654, Cambridge, MA,
  USA, 2014.

\bibitem[Duan et~al.(2016)Duan, Chen, Houthooft, Schulman, and
  Abbeel]{Duan2016}
Y.~Duan, X.~Chen, R.~Houthooft, J.~Schulman, and P.~Abbeel.
\newblock Benchmarking deep reinforcement learning for continuous control.
\newblock In \emph{Proceedings of the 33rd International Conference on
  International Conference on Machine Learning - Volume 48}, pages 1329--1338,
  2016.

\bibitem[Fang et~al.(2018)Fang, Li, Lin, and Zhang]{Fang2018}
C.~Fang, C.~J. Li, Z.~Lin, and T.~Zhang.
\newblock Spider: Near-optimal non-convex optimization via stochastic
  path-integrated differential estimator.
\newblock In \emph{NeurIPS}, 2018.

\bibitem[Haarnoja et~al.(2018)Haarnoja, Zhou, Abbeel, and Levine]{Haarnoja2018}
T.~Haarnoja, A.~Zhou, P.~Abbeel, and S.~Levine.
\newblock Soft actor-critic: Off-policy maximum entropy deep reinforcement
  learning with a stochastic actor.
\newblock In \emph{Proceedings of the 35th International Conference on Machine
  Learning}, volume~80 of \emph{Proceedings of Machine Learning Research},
  pages 1861--1870, Stockholmsmässan, Stockholm Sweden, 10--15 Jul 2018. PMLR.

\bibitem[Hasselt et~al.(2016)Hasselt, Guez, and Silver]{Hasselt2016}
H.~van Hasselt, A.~Guez, and D.~Silver.
\newblock Deep reinforcement learning with double q-learning.
\newblock In \emph{Proceedings of the Thirtieth AAAI Conference on Artificial
  Intelligence}, pages 2094--2100, 2016.

\bibitem[Johnson and Zhang(2013)]{johnson2013accelerating}
R.~Johnson and T.~Zhang.
\newblock Accelerating stochastic gradient descent using predictive variance
  reduction.
\newblock In \emph{Advances in Neural Information Processing Systems (NIPS)},
  pages 315--323, 2013.

\bibitem[Kingma and Ba(2014)]{Kingma2014}
D.~P. Kingma and J.~Ba.
\newblock Adam: A method for stochastic optimization.
\newblock \emph{CoRR}, abs/1412.6980, 2014.

\bibitem[Klimov and Schulman(2017)]{roboschool}
O.~Klimov and J.~Schulman.
\newblock {Roboschool}.
\newblock
  \href{https://openai.com/blog/roboschool/}{https://openai.com/blog/roboschool/},
  2017.

\bibitem[Levine(2017)]{BerkeleyRL}
S.~Levine.
\newblock {CS 294-112}: Deep reinforcement learning lecture notes, 2017.

\bibitem[Li and Li(2018)]{Li2018}
Z.~Li and J.~Li.
\newblock A simple proximal stochastic gradient method for nonsmooth nonconvex
  optimization.
\newblock In \emph{Proceedings of the 32Nd International Conference on Neural
  Information Processing Systems}, pages 5569--5579, USA, 2018.

\bibitem[Lillicrap et~al.(2016)Lillicrap, Hunt, Pritzel, Heess, Erez, Tassa,
  Silver, and Wierstra]{Lillicrap2016}
T.~P. Lillicrap, J.~J. Hunt, A.~Pritzel, N.~Heess, T.~Erez, Y.~Tassa,
  D.~Silver, and D.~Wierstra.
\newblock Continuous control with deep reinforcement learning.
\newblock In \emph{4th International Conference on Learning Representations,
  {ICLR} 2016, San Juan, Puerto Rico, May 2-4, 2016, Conference Track
  Proceedings}, 2016.

\bibitem[Liu et~al.(2019)Liu, Li, Kang, and Darrell]{liu2019regularization}
Z.~Liu, X.~Li, B.~Kang, and T.~Darrell.
\newblock Regularization matters in policy optimization.
\newblock \emph{arXiv preprint arXiv:1910.09191}, 2019.

\bibitem[Mnih et~al.(2013)Mnih, Kavukcuoglu, Silver, Graves, Antonoglou,
  Wierstra, and Riedmiller]{Mnih2013}
V.~Mnih, K.~Kavukcuoglu, D.~Silver, A.~Graves, I.~Antonoglou, D.~Wierstra, and
  M.~A. Riedmiller.
\newblock Playing atari with deep reinforcement learning.
\newblock \emph{ArXiv}, abs/1312.5602, 2013.

\bibitem[Mnih et~al.(2015)Mnih, Kavukcuoglu, Silver, Rusu, Veness, Bellemare,
  Graves, Riedmiller, Fidjeland, Ostrovski, Petersen, Beattie, Sadik,
  Antonoglou, King, Kumaran, Wierstra, Legg, and Hassabis]{Mnih2015}
V.~Mnih, K.~Kavukcuoglu, D.~Silver, A.~A. Rusu, J.~Veness, M.~G. Bellemare,
  A.~Graves, M.~A. Riedmiller, A.~Fidjeland, G.~Ostrovski, S.~Petersen,
  C.~Beattie, A.~Sadik, I.~Antonoglou, H.~King, D.~Kumaran, D.~Wierstra,
  S.~Legg, and D.~Hassabis.
\newblock Human-level control through deep reinforcement learning.
\newblock \emph{Nature}, 518:\penalty0 529--533, 2015.

\bibitem[Mnih et~al.(2016)Mnih, Badia, Mirza, Graves, Lillicrap, Harley,
  Silver, and Kavukcuoglu]{Mnih2016}
V.~Mnih, A.~P. Badia, M.~Mirza, A.~Graves, Timothy Lillicrap, Tim Harley, David
  Silver, and Koray Kavukcuoglu.
\newblock Asynchronous methods for deep reinforcement learning.
\newblock In \emph{Proceedings of The 33rd International Conference on Machine
  Learning}, volume~48, pages 1928--1937, 20--22 Jun 2016.

\bibitem[Nesterov(2014)]{Nesterov2014}
Y.~Nesterov.
\newblock \emph{Introductory Lectures on Convex Optimization: A Basic Course}.
\newblock Springer Publishing Company, Incorporated, 1 edition, 2014.

\bibitem[Neyshabur et~al.(2017)Neyshabur, Bhojanapalli, Mcallester, and
  Srebro]{Neyshabur2017}
B.~Neyshabur, S.~Bhojanapalli, D.~Mcallester, and N.~Srebro.
\newblock Exploring generalization in deep learning.
\newblock In \emph{Advances in Neural Information Processing Systems 30}, pages
  5947--5956. 2017.

\bibitem[Nguyen et~al.(2017{\natexlab{a}})Nguyen, Liu, Scheinberg, and
  M.Tak{\'{a}}c]{Nguyen2017}
L.~M. Nguyen, J.~Liu, K.~Scheinberg, and M.Tak{\'{a}}c.
\newblock Stochastic recursive gradient algorithm for nonconvex optimization.
\newblock \emph{CoRR}, abs/1705.07261, 2017{\natexlab{a}}.

\bibitem[Nguyen et~al.(2017{\natexlab{b}})Nguyen, Liu, Scheinberg, and
  Tak{\'a}{\v{c}}]{Nguyen2017b}
L.~M. Nguyen, J.~Liu, K.~Scheinberg, and M.~Tak{\'a}{\v{c}}.
\newblock {SARAH}: A novel method for machine learning problems using
  stochastic recursive gradient.
\newblock In \emph{Proceedings of the 34th International Conference on Machine
  Learning}, pages 2613--2621, 2017{\natexlab{b}}.

\bibitem[Nguyen et~al.(2019)Nguyen, van Dijk, Phan, Nguyen, Weng, and
  Kalagnanam]{Nguyen2019_SARAH}
L.~M. Nguyen, M.~van Dijk, D.~T. Phan, P.~H. Nguyen, T.-W. Weng, and J.~R.
  Kalagnanam.
\newblock Finite-sum smooth optimization with sarah.
\newblock \emph{arXiv preprint arXiv:1901.07648}, 2019.

\bibitem[OpenAI(2018)]{OpenAI_dota}
OpenAI.
\newblock {OpenAI Five}.
\newblock \url{https://blog.openai.com/openai-five/}, 2018.

\bibitem[Papini et~al.(2018)Papini, Binaghi, Canonaco, Pirotta, and
  Restelli]{Papini2018a}
M.~Papini, D.~Binaghi, G.~Canonaco, M.~Pirotta, and M.~Restelli.
\newblock Stochastic variance-reduced policy gradient.
\newblock In \emph{Proceedings of the 35th International Conference on Machine
  Learning}, volume~80, pages 4026--4035, 10--15 Jul 2018.

\bibitem[Parikh and Boyd(2014)]{Parikh2014}
N.~Parikh and S.~Boyd.
\newblock Proximal algorithms.
\newblock \emph{Found. Trends Optim.}, 1\penalty0 (3):\penalty0 127--239,
  January 2014.

\bibitem[Pham et~al.(2019{\natexlab{b}})Pham, Nguyen, Phan, and
  Tran-Dinh]{Pham2019ProxSARAH}
N.~H. Pham, L.~M. Nguyen, D.~T. Phan, and Q.~Tran-Dinh.
\newblock Proxsarah: An efficient algorithmic framework for stochastic
  composite nonconvex optimization.
\newblock \emph{ArXiv}, abs/1902.05679, 2019{\natexlab{b}}.

\bibitem[Reddi et~al.(2016)Reddi, Hefny, Sra, P\'{o}cz\'{o}s, and
  Smola]{Reddi2016}
S.~J. Reddi, A.~Hefny, S.~Sra, B.~P\'{o}cz\'{o}s, and A.~Smola.
\newblock Stochastic variance reduction for nonconvex optimization.
\newblock In \emph{Proceedings of the 33rd International Conference on
  International Conference on Machine Learning - Volume 48}, pages 314--323,
  2016.

\bibitem[Robbins and Monro(1951)]{Robbins1951}
H.~Robbins and S.~Monro.
\newblock A stochastic approximation method.
\newblock \emph{Ann. Math. Statist.}, 22\penalty0 (3):\penalty0 400--407, 09
  1951.

\bibitem[Schulman et~al.(2015)Schulman, Levine, Abbeel, Jordan, and
  Moritz]{Schulman2015}
J.~Schulman, S.~Levine, P.~Abbeel, Michael Jordan, and Philipp Moritz.
\newblock Trust region policy optimization.
\newblock In \emph{Proceedings of the 32nd International Conference on Machine
  Learning}, volume~37, pages 1889--1897, Lille, France, 07--09 Jul 2015.

\bibitem[Schulman et~al.(2017)Schulman, Wolski, Dhariwal, Radford, and
  Klimov]{Schulman2017}
J.~Schulman, F.~Wolski, P.~Dhariwal, A.~Radford, and O.~Klimov.
\newblock Proximal policy optimization algorithms.
\newblock \emph{ArXiv}, abs/1707.06347, 2017.

\bibitem[Shen et~al.(2019)Shen, Ribeiro, Hassani, Qian, and Mi]{Shen2019d}
Z.~Shen, A.~Ribeiro, H.~Hassani, H.~Qian, and C.~Mi.
\newblock Hessian aided policy gradient.
\newblock In \emph{Proceedings of the 36th International Conference on Machine
  Learning}, volume~97 of \emph{Proceedings of Machine Learning Research},
  pages 5729--5738, Long Beach, California, USA, 09--15 Jun 2019.

\bibitem[Silver et~al.(2014)Silver, Lever, Heess, Degris, Wierstra, and
  Riedmiller]{Silver2014}
D.~Silver, G.~Lever, N.~Heess, T.~Degris, D.~Wierstra, and M.~Riedmiller.
\newblock Deterministic policy gradient algorithms.
\newblock In \emph{Proceedings of the 31st International Conference on
  International Conference on Machine Learning - Volume 32}, pages
  I--387--I--395, 2014.

\bibitem[Silver et~al.(2016)Silver, Huang, Maddison, Guez, Sifre, v.~d.
  Driessche, Schrittwieser, Antonoglou, Panneershelvam, Lanctot, Dieleman,
  Grewe, Nham, Kalchbrenner, Sutskever, Lillicrap, Leach, Kavukcuoglu, Graepel,
  and Hassabis]{Silver2016}
D.~Silver, A.~Huang, C.~J. Maddison, A.~Guez, L.~Sifre, G.~v.~d. Driessche,
  J.~Schrittwieser, I.~Antonoglou, V.~Panneershelvam, M.~Lanctot, S.~Dieleman,
  D.~Grewe, J.~Nham, N.~Kalchbrenner, I.~Sutskever, T.~Lillicrap, M.~Leach,
  K.~Kavukcuoglu, T.~Graepel, and D.~Hassabis.
\newblock Mastering the game of go with deep neural networks and tree search.
\newblock \emph{Nature}, 529:\penalty0 484--503, 2016.

\bibitem[Silver et~al.(2018)Silver, Hubert, Schrittwieser, Antonoglou, Lai,
  Guez, Lanctot, Sifre, Kumaran, Graepel, Lillicrap, Simonyan, and
  Hassabis]{Silver2018}
D.~Silver, T.~Hubert, J.~Schrittwieser, I.~Antonoglou, M.~Lai, A.~Guez,
  M.~Lanctot, L.~Sifre, D.~Kumaran, T.~Graepel, T.~Lillicrap, K.~Simonyan, and
  D.~Hassabis.
\newblock A general reinforcement learning algorithm that masters chess, shogi,
  and go through self-play.
\newblock \emph{Science}, 362\penalty0 (6419):\penalty0 1140--1144, 2018.

\bibitem[Srivastava et~al.(2014)Srivastava, Hinton, Krizhevsky, Sutskever, and
  Salakhutdinov]{srivastava2014dropout}
N.~Srivastava, G.~Hinton, A.~Krizhevsky, I.~Sutskever, and R.~Salakhutdinov.
\newblock Dropout: a simple way to prevent neural networks from overfitting.
\newblock \emph{The journal of machine learning research}, 15\penalty0
  (1):\penalty0 1929--1958, 2014.

\bibitem[Sutton and Barto(2018)]{Sutton2018}
R.~S. Sutton and A.~G. Barto.
\newblock \emph{Introduction to Reinforcement Learning, 2nd Edition}.
\newblock MIT Press, 2018.

\bibitem[Sutton et~al.(1999)Sutton, McAllester, Singh, and Mansour]{Sutton1999}
R.~S. Sutton, D.~McAllester, S.~Singh, and Y.~Mansour.
\newblock Policy gradient methods for reinforcement learning with function
  approximation.
\newblock In \emph{Proceedings of the 12th International Conference on Neural
  Information Processing Systems}, pages 1057--1063, 1999.

\bibitem[Tieleman and Hinton(2012)]{Tieleman2012}
T.~Tieleman and G.~Hinton.
\newblock {Lecture 6.5---RmsProp: Divide the gradient by a running average of
  its recent magnitude}.
\newblock COURSERA: Neural Networks for Machine Learning, 2012.

\bibitem[Todorov et~al.(2012)Todorov, Erez, and Tassa]{Todorov2012MuJoCoAP}
E.~Todorov, T.~Erez, and Y.~Tassa.
\newblock Mujoco: A physics engine for model-based control.
\newblock \emph{2012 IEEE/RSJ International Conference on Intelligent Robots
  and Systems}, pages 5026--5033, 2012.

\bibitem[Tran-Dinh et~al.(2019{\natexlab{a}})Tran-Dinh, Pham, Phan, and
  Nguyen]{tran2019hybrid1}
Q.~Tran-Dinh, N.~H. Pham, D.~T. Phan, and L.~M. Nguyen.
\newblock Hybrid stochastic gradient descent algorithms for stochastic
  nonconvex optimization.
\newblock \emph{arXiv preprint arXiv:1905.05920}, 2019{\natexlab{a}}.

\bibitem[Tran-Dinh et~al.(2019{\natexlab{b}})Tran-Dinh, Pham, Phan, and
  Nguyen]{tran2019hybrid2}
Q.~Tran-Dinh, N.~H. Pham, D.~T. Phan, and L.~M. Nguyen.
\newblock A hybrid stochastic optimization framework for stochastic composite
  nonconvex optimization.
\newblock \emph{arXiv preprint arXiv:1907.03793}, 2019{\natexlab{b}}.

\bibitem[Wang et~al.(2016)Wang, Schaul, Hessel, Van~Hasselt, Lanctot, and
  De~Freitas]{Wang2016}
Z.~Wang, T.~Schaul, M.~Hessel, H.~Van~Hasselt, M.~Lanctot, and N.~De~Freitas.
\newblock Dueling network architectures for deep reinforcement learning.
\newblock In \emph{Proceedings of the 33rd International Conference on
  International Conference on Machine Learning - Volume 48}, pages 1995--2003,
  2016.

\bibitem[Wang et~al.(2017)Wang, Bapst, Heess, Mnih, Munos, Kavukcuoglu, and
  d.~Freitas]{Wang2017}
Z.~Wang, V.~Bapst, N.~Heess, V.~Mnih, R.~Munos, K.~Kavukcuoglu, and
  N.~d.~Freitas.
\newblock Sample efficient actor-critic with experience replay.
\newblock In \emph{5th International Conference on Learning Representations,
  {ICLR} 2017, Toulon, France, April 24-26, 2017, Conference Track
  Proceedings}, 2017.

\bibitem[Wang et~al.(2018)Wang, Ji, Zhou, Liang, and Tarokh]{Wang2018}
Z.~Wang, K.~Ji, Y.~Zhou, Y.~Liang, and V.~Tarokh.
\newblock Spiderboost: A class of faster variance-reduced algorithms for
  nonconvex optimization.
\newblock \emph{ArXiv}, abs/1810.10690, 2018.

\bibitem[Watkins and Dayan(1992)]{Watkins1992}
C.~J. C.~H. Watkins and P.~Dayan.
\newblock Q-learning.
\newblock \emph{Machine Learning}, 8\penalty0 (3):\penalty0 279--292, May 1992.

\bibitem[Williams(1992)]{Williams1992}
R.~J. Williams.
\newblock Simple statistical gradient-following algorithms for connectionist
  reinforcement learning.
\newblock \emph{Machine Learning}, 8\penalty0 (3):\penalty0 229--256, May 1992.

\bibitem[Wu et~al.(2017)Wu, Mansimov, Liao, Grosse, and Ba]{Wu2017}
Y.~Wu, E.~Mansimov, S.~Liao, R.~Grosse, and J.~Ba.
\newblock Scalable trust-region method for deep reinforcement learning using
  kronecker-factored approximation.
\newblock In \emph{Proceedings of the 31st International Conference on Neural
  Information Processing Systems}, NIPS'17, pages 5285--5294, USA, 2017.

\bibitem[Xu et~al.(2019{\natexlab{a}})Xu, Gao, and Gu]{Xu2019}
P.~Xu, F.~Gao, and Q.~Gu.
\newblock An improved convergence analysis of stochastic variance-reduced
  policy gradient.
\newblock \emph{Conference on Uncertainty in Artificial Intelligence},
  2019{\natexlab{a}}.

\bibitem[Xu et~al.(2019{\natexlab{b}})Xu, Gao, and Gu]{Xu2019SampleEP}
P.~Xu, F.~Gao, and Q.~Gu.
\newblock Sample efficient policy gradient methods with recursive variance
  reduction.
\newblock \emph{ArXiv}, abs/1909.08610, 2019{\natexlab{b}}.

\bibitem[Yang and Zhang(2019)]{Yang2019}
L.~Yang and Y.~Zhang.
\newblock Policy optimization with stochastic mirror descent.
\newblock \emph{CoRR}, abs/1906.10462, 2019.

\bibitem[Yuan et~al.(2019)Yuan, Li, Tang, and Zhou]{yuan2019policy}
H.~Yuan, C.~J. Li, Y.~Tang, and Y.~Zhou.
\newblock Policy optimization via stochastic recursive gradient algorithm,
  2019.
\newblock URL \url{https://openreview.net/forum?id=rJl3S2A9t7}.

\bibitem[Zhang et~al.(2017)Zhang, Bengio, Hardt, Recht, and Vinyals]{Zhang2017}
C.~Zhang, S.~Bengio, M.~Hardt, B.~Recht, and O.~Vinyals.
\newblock Understanding deep learning requires rethinking generalization.
\newblock 2017.
\newblock URL \url{https://arxiv.org/abs/1611.03530}.

\bibitem[Zhao et~al.(2011)Zhao, Hachiya, Niu, and Sugiyama]{Zhao2011}
T.~Zhao, H.~Hachiya, G.~Niu, and M.~Sugiyama.
\newblock Analysis and improvement of policy gradient estimation.
\newblock In \emph{Advances in Neural Information Processing Systems 24}, pages
  262--270. 2011.

\bibitem[Zhou et~al.(2018)Zhou, Xu, and Gu]{Zhou2018StochasticNV}
D.~Zhou, P.~Xu, and Q.~Gu.
\newblock Stochastic nested variance reduction for nonconvex optimization.
\newblock \emph{ArXiv}, abs/1806.07811, 2018.

\end{thebibliography}


\clearpage
\newpage
\onecolumn
\appendix
\begin{center}
\textsc{\large Supplementary document}

\textbf{\Large A Hybrid Stochastic Policy Gradient Algorithm for Reinforcement Learning}
\end{center}
This supplementary document presents the full proofs of technical results presented in the main text.
It also provides the details of our configurations for numerical experiments in Section~\ref{sec:experiments}.

\section{Convergence Analysis}\label{sec:appendix1}
We note that the original idea of using hybrid estimators has been proposed in our working paper \citep{tran2019hybrid2}. 
In this work, we have extended this idea as well as the proof techniques for stochastic optimization in \cite{tran2019hybrid2} into reinforcement learning settings. 
We now provide the full analysis of Algorithm~\ref{alg:A1} and \ref{alg:A2}.
We first prove a key property of our new hybrid estimator for the policy gradient $\nabla{J}(\theta)$.
Then, we provide the proof of Theorem~\ref{thm:grad_bound} and Corollary \ref{cor:complexity}. 

\subsection{Proof of Lemma~\ref{lem:prop_hybrid}: Bound on the Variance of the Hybrid SPG Estimator} \label{app:A1}
Part of this proof comes from the proof of Lemma 1 in \cite{tran2019hybrid2}. Let $\Exps{\Bc,\widehat{\Bc}}{\cdot} := \Exps{\tau,\hat{\tau}\sim p_{\theta_t}}{\cdot}$ be the total expectation. Using the independence of $\tau$ and $\hat{\tau}$, taking the total expectation on \eqref{eq:hybrid_est}, we obtain
\begin{equation*}
\begin{array}{ll}
\Exps{\Bc,\widehat{\Bc}}{v_t} &= \beta v_{t-1} + \beta \left[ \nabla J(\theta_t) - \nabla J(\theta_{t-1}) \right] + (1-\beta)\nabla J(\theta_t) \vspace{1ex}\\
&= \nabla J(\theta_t) + \beta\left[ v_{t-1} - \nabla J(\theta_{t-1}\right],
\end{array}
\end{equation*}
which is the same as \eqref{eq:lem41_1}.

To prove \eqref{eq:lem41_2}, we first define $u_t := \frac{1}{B}\dsum_{\hat{\tau}\in \widehat{\Bc}_t}g(\hat{\tau}\vert\theta_t)$ and $\Delta u_t := u_t - \nabla J(\theta_t)$.
We have
\begin{equation*}
\begin{array}{ll}
     \|\Delta v_t\|^2 &= \beta^2\|\Delta v_{t-1}\|^2 + \frac{\beta^2}{B^2}\norm{\displaystyle\sum_{\tau\in \Bc_t}\Delta g(\tau\vert\theta_t)}^2 + (1-\beta)^2\|\Delta u_t\|^2 + \beta^2\|\nabla J(\theta_{t-1}) - \nabla J(\theta_t)\|^2 \vspace{1ex}\\
     &+ \frac{2\beta^2}{B}\dsum_{\tau\in \Bc_t}(\Delta v_{t-1})^{\top}[\Delta g(\tau\vert\theta_t)] + 2\beta^2(\Delta v_{t-1})^{\top}[\nabla J(\theta_{t-1}) - \nabla J(\theta_t)] \vspace{1ex}\\
     &+ 2\beta(1-\beta)(\Delta v_{t-1})^{\top}[u_t - \nabla J(\theta_t)] + \frac{2\beta(1-\beta)}{B}\dsum_{\tau\in \Bc_t}[\Delta g(\tau\vert\theta_t)]^{\top}(\Delta u_t) \vspace{1ex}\\
     &+  \frac{2\beta^2}{B}\dsum_{\tau\in \Bc_t}(\Delta g(\tau\vert\theta_t))^{\top}[\nabla J(\theta_{t-1}) - \nabla J(\theta_t)] + 2\beta(1-\beta)(\Delta u_t)^{\top}[\nabla J(\theta_{t-1}) - \nabla J(\theta_t)].
\end{array}
\end{equation*}
Taking the total expectation and note that $\Exps{\widehat{\Bc}}{u_t} := \Exps{\hat{\tau} \sim p_{\theta_t}}{u_t} = \nabla J(\theta_t)$ and $\Exps{\widehat{\Bc}}{\|u_t - \nabla J(\theta_t)\|^2} \le \frac{1}{B^2}\dsum_{\hat{\tau}\in\widehat{\Bc}}\Exp{\|g(\hat{\tau}|\theta_t) - \Exp{g(\hat{\tau}|\theta_t)}\|^2} = \frac{\sigma^2}{B}$, we get
\begin{equation}\label{eq:lem42_eq1}
    \begin{array}{ll}
     \Exps{\Bc,\widehat{\Bc}}{\|\Delta v_t\|^2 } 
&= \beta^2\|\Delta v_{t-1}\|^2 + \frac{\beta^2}{B^2}\Exps{\Bc}{\Big\Vert \dsum_{\tau \in \Bc_t}\Delta g(\tau \vert\theta_t)\Big\Vert^2 }+(1-\beta)^2\Exps{\widehat{\Bc}}{\norms{\Delta u_t}^2} \vspace{1ex}\\
&- \beta^2\|\nabla J(\theta_{t-1}) - \nabla J(\theta_t)\|^2 \vspace{1ex}\\
     &\le \beta^2\| \Delta v_{t-1}\|^2 + \frac{\beta^2}{B^2}\dsum_{\tau \in \Bc_t}\Exps{\Bc}{\|\Delta g(\tau \vert \theta_t)\|^2} - \beta^2\|\nabla J(\theta_{t-1}) - \nabla J(\theta_t)\|^2 \vspace{1ex}\\
     &+ {~} \frac{(1-\beta)^2\sigma^2}{B} \vspace{1ex}\\
     &\le \beta^2\| \Delta v_{t-1}\|^2 + \frac{\beta^2}{B^2}\dsum_{\tau \in \Bc_t}\Exps{\Bc}{\|\Delta g(\tau \vert \theta_t)\|^2} + \frac{(1-\beta)^2}{B}\sigma^2,
\end{array}
\end{equation}
where the first inequality comes from the triangle inequality then we ignore the non-negative terms to arrive at the second inequality.

Additionally, Lemma 6.1 in \cite{Xu2019} shows that
\begin{equation}\label{eq:xu_lem}
\Var{\omega(\tau\vert \theta_t, \theta_{t-1})} \le C_\omega \norm{\theta_t - \theta_{t-1}}^2,
\end{equation}
where $C_\omega := H(2HG^2 + M)(W + 1)$.

Using \eqref{eq:xu_lem} we have
\begin{equation*}
\begin{array}{ll}
\Exps{\Bc}{\norm{\Delta g(\tau \vert \theta_t)}^2} &= \Exps{\Bc}{\norm{g(\tau\vert\theta_t) - \omega(\tau \vert\theta_t,\theta_{t-1}) g(\tau \vert\theta_{t-1})}^2} \vspace{1ex}\\
&= \Exps{\Bc}{\norm{[1 - \omega(\tau \vert\theta_t,\theta_{t-1})]g(\tau\vert\theta_{t-1}) + (g(\tau\vert\theta_t) - g(\tau\vert\theta_{t-1})}^2} \vspace{1ex}\\
&\le \Exps{\Bc}{\norm{[1 - \omega(\tau \vert\theta_t,\theta_{t-1})]g(\tau\vert\theta_{t-1})}^2} + \Exps{\Bc}{\norm{g(\tau\vert\theta_t) - g(\tau\vert\theta_{t-1})}^2} \vspace{1ex}\\
&\overset{(\star)}{\le} C_g^2\Exps{\Bc}{\norm{1 - \omega(\tau \vert\theta_t,\theta_{t-1})}^2} + L_g^2\norm{\theta_t - \theta_{t-1}}^2 \vspace{1ex}\\
&\overset{(\star\star)}{=} C_g^2 \Var{\omega(\tau\vert\theta_t,\theta_{t-1})} + L_g^2\norm{\theta_t - \theta_{t-1}}^2 \vspace{1ex}\\
&\overset{\eqref{eq:xu_lem}}{\le} \left(C_g^2 C_\omega + L_g^2 \right) \norm{\theta_t - \theta_{t-1}}^2,
\end{array}
\end{equation*}
where $L_g := \frac{HM (R + \abs{b})}{(1-\gamma)}$, $C_g := \frac{HG(R + \abs{b})}{(1-\gamma)}$, and $b$ is a baseline reward. Here, $(\star)$ comes from Lemma~\ref{lem:smoothness} and $(\star\star)$ is from Lemma~1 in \cite{Cortes2010}.

Plugging the last estimate into \eqref{eq:lem42_eq1} yields
\begin{equation}\label{eq:lem42_eq2}
\begin{array}{ll}
 \Exps{\Bc,\widehat{\Bc}}{\|\Delta v_t\|^2 } &\le \beta^2\| \Delta v_{t-1}\|^2 + \dfrac{\beta^2(C_g^2C_\omega + L_g^2)}{B}\norm{\theta_t - \theta_{t-1}}^2  +\dfrac{(1-\beta)^2}{B}\sigma^2,
 \end{array}
\end{equation}
which is \eqref{eq:lem41_2}, where $\overline{C} := C_g^2C_\omega + L_g^2$.
\Eproof

\subsection{Proof of Lemma~\ref{lem:key_estimate}: Key Estimate of Algorithm~\ref{alg:A1}}
Similar to the proof of Lemma 5 in \cite{tran2019hybrid2} , from the update in Algorithm~\ref{alg:A1}, we have $\theta_{t+1} = (1-\gamma)\theta_t + \gamma \widehat{\theta}_{t+1}$, which leads to $\theta_{t+1} - \theta_t = \gamma(\widehat{\theta}_{t+1} - \theta_t)$. 
Combining this expression and the $L$-smoothness of $J(\theta)$ in Lemma~\ref{lem:smoothness}, we have
\begin{equation}\label{eq:lem43_eq1}
    \begin{array}{ll}
        J(\theta_{t+1}) &\ge  J(\theta_t) + \left[\nabla J(\theta_t)\right]^{\top}(\theta_{t+1} - \theta_t) - \frac{L}{2}\|\theta_{t+1} - \theta_t\|^2 \vspace{1ex}\\
 & = J(\theta_t) + \alpha\left[\nabla J(\theta_t)\right]^{\top}(\widehat{\theta}_{t+1} - \theta_t) - \frac{L\alpha^2}{2}\norms{\widehat{\theta}_{t+1} - \theta_t}^2.
    \end{array}
\end{equation}
From the convexity of $Q$, we have
\begin{equation}\label{eq:lem43_eq2}
Q(\theta_{t+1}) \le (1-\alpha)Q(\theta_t) + \alpha Q(\widehat{\theta}_{t+1}) \le Q(\theta_t) + \alpha \nabla Q(\widehat{\theta}_{t+1})^{\top}(\widehat{\theta}_{t+1} - \theta_t),
\end{equation}
where $\nabla Q(\widehat{\theta}_{t+1})$ is a subgradient of $Q$ at $\widehat{\theta}_{t+1}$.

By the optimality condition of $\widehat{\theta}_{t+1} = \prox_{\eta Q}(\theta_t + \eta v_t)$, we can show that $\nabla Q(\widehat{\theta}_{t+1}) = v_t - \frac{1}{\eta}(\widehat{\theta}_{t+1} - \theta_t)$ for some $\nabla Q(\widehat{\theta}_{t+1}) \in \partial Q(\widehat{\theta}_{t+1})$ where $\partial Q$ is the subdifferential of Q at $\widehat{\theta}_{t+1}$. Plugging this into \eqref{eq:lem43_eq2}, we get
\begin{equation}\label{eq:lem43_eq3}
Q(\theta_{t+1}) \le Q(\theta_t) + \alpha v_t^{\top}(\widehat{\theta}_{t+1} - \theta_t) - \frac{\alpha}{\eta}\norms{\widehat{\theta}_{t+1} - \theta_t}^2.
\end{equation}
Subtracting \eqref{eq:lem43_eq3} from \eqref{eq:lem43_eq1}, we obtain
\begin{equation}\label{eq:lem43_eq4}
\begin{array}{ll}
F(\theta_{t+1}) &\ge F(\theta_t) + \alpha\left[\nabla J(\theta_t) - v_t \right]^{\top}(\widehat{\theta}_{t+1} - \theta_t) + \left( \frac{\alpha}{\eta} - \frac{L\alpha^2}{2} \right) \norms{\widehat{\theta}_{t+1} - \theta_t}^2\vspace{1ex}\\
&= F(\theta_t) - \alpha\left[v_t-\nabla J(\theta_t) \right]^{\top}(\widehat{\theta}_{t+1} - \theta_t) + \left( \frac{\alpha}{\eta} - \frac{L\alpha^2}{2} \right) \norms{\widehat{\theta}_{t+1} - \theta_t}.
\end{array}
\end{equation}
Using the fact that
\begin{equation*}
\begin{array}{ll}
\left[v_t-\nabla J(\theta_t)\right]^{\top}(\widehat{\theta}_{t+1} - \theta_t) &= \frac{1}{2}\norm{v_t-\nabla J(\theta_t) }^2 + \frac{1}{2}\norms{\widehat{\theta}_{t+1}- \theta_t}^2   \vspace{1ex}\\
&- \frac{1}{2}\norms{v_t-\nabla J(\theta_t) - (\widehat{\theta}_{t+1} - \theta_t)}^2,
\end{array}
\end{equation*}
and ignoring the non-negative term $\frac{1}{2}\norms{v_t-\nabla J(\theta_t) - (\widehat{\theta}_{t+1} - \theta_t)}^2$, we can rewrite \eqref{eq:lem43_eq4} as
\begin{equation*}
F(\theta_{t+1}) \ge F(\theta_t) - \dfrac{\alpha}{2}\norm{\nabla J(\theta_t) - v_t}^2  + \left( \dfrac{\alpha}{\eta} - \dfrac{L\alpha^2}{2} - \frac{\alpha }{2} \right) \norms{\widehat{\theta}_{t+1} - \theta_t}^2.
\end{equation*}
Taking the total expectation over the entire history $\Fc_{t+1}$, we obtain
\begin{equation}\label{eq:lem43_eq5}
\Exp{F(\theta_{t+1})} \ge \Exp{F(\theta_t)} - \dfrac{\alpha}{2}\Exp{\norm{\nabla J(\theta_t) - v_t}^2}  + \left( \dfrac{\alpha}{\eta} - \dfrac{L\alpha^2}{2} - \frac{\alpha}{2} \right) \Exp{\norms{\widehat{\theta}_{t+1} - \theta_t}^2}.
\end{equation}

From the definition of the gradient mapping \eqref{eq:grad_map_def}, we have
\begin{equation*}
\eta \norms{\Gc_\eta(\theta_t) } = \norms{\prox_{\eta Q}(\theta_t + \eta \nabla J(\theta_t)) - \theta_t}.
\end{equation*}
Applying the triangle inequality, we can derive
\begin{equation*}
\begin{array}{ll}
\eta \norm{\Gc_\eta(\theta_t) } &\le  \norms{\widehat{\theta}_{t+1} -  \theta_t} +\norms{ \prox_{\eta Q}(\theta_t + \eta \nabla J(\theta_t)) - \widehat{\theta}_{t+1}} \vspace{1ex}\\
&= \norms{\widehat{\theta}_{t+1} -  \theta_t} +\norms{\prox_{\eta Q}(\theta_t + \eta \nabla J(\theta_t)) - \prox_{\eta Q}(\theta_t + \eta v_t)} \vspace{1ex}\\
&\le \norms{\widehat{\theta}_{t+1} -  \theta_t} +\eta\norms{ v_{t} - \nabla J(\theta_{t})}.
\end{array}
\end{equation*}

Taking the full expectation over the entire history $\Fc_{t+1}$ yields
\begin{equation*}
\eta^2\Exp{\Gc_\eta(\theta_t)}^2 \le 2\Exp{\norms{\widehat{\theta}_{t+1} - \theta_t}^2} + 2\eta^2\Exp{\norms{v_{t} - \nabla J(\theta_{t})}^2}.
\end{equation*}
Multiply this inequality by $-\frac{\alpha}{2}$ and add to \eqref{eq:lem43_eq5}, we arrive at
\begin{equation*}
\begin{array}{ll}
\Exp{F(\theta_{t+1})}  &\ge \Exp{F(\theta_t) }+ \frac{\eta^2 \alpha}{2} \Exp{\norms{\Gc_\eta(\theta_t)}^2} - \frac{\alpha}{2}\left(1 + 2\eta^2 \right)\Exp{\norms{v_{t} - \nabla J(\theta_{t})}^2} \vspace{1ex}\\
& + {~} \frac{\alpha}{2}\left(\frac{2}{\eta} - L\alpha -  3 \right) \Exp{\norms{\widehat{\theta}_{t+1} - \theta_t}^2},
\end{array}
\end{equation*}
which can be rewritten as
\begin{equation*}
\begin{array}{l}
\Exp{F(\theta_{t+1})}  \ge \Exp{F(\theta_t) }+ \frac{\eta^2 \alpha}{2} \Exp{\norms{\Gc_\eta(\theta_t)}^2} - \frac{\xi}{2}\Exp{\norms{v_{t} - \nabla J(\theta_{t})}^2} + \frac{\zeta}{2}\Exp{\norms{\widehat{\theta}_{t+1} - \theta_t}^2},
\end{array}
\end{equation*}
where $\xi := \alpha(1 + 2\eta^2)$ and $\zeta := \alpha\left(\frac{2}{\eta} - L\alpha - 3\right)$ which is exactly \eqref{eq:lem43}.
\Eproof

\subsection{Proof of Theorem~\ref{thm:grad_bound}: Key Bound on the Gradient Mapping}\label{app:A3}
Firstly, using the identity $\theta_{t+1} - \theta_t = \gamma(\widehat{\theta}_{t+1} - \theta_t)$, taking the total expectation over the entire history $\Fc_{t+1}$, we can rewrite \eqref{eq:lem41_2} as
\begin{equation}\label{eq:thm_eq1}
\begin{array}{ll}
\Exp{\norm{v_{t+1} - \nabla J(\theta_{t+1})}^2} &\le \beta^2 \Exp{\norm{v_{t} - \nabla J(\theta_{t})}^2} + \frac{\beta^2\overline{C}}{B}\Exp{\norms{\theta_{t+1} - \theta_{t}}^2} + \frac{(1-\beta)^2}{B}\sigma^2 \vspace{1ex}\\
&= \beta^2 \Exp{\norms{v_{t} - \nabla J(\theta_{t})}^2} + \frac{\beta^2\overline{C}\alpha^2}{B}\Exp{\norms{\widehat{\theta}_{t+1} - \theta_{t}}^2} + \frac{(1-\beta)^2}{B}\sigma^2.
\end{array}
\end{equation}

Multiply \eqref{eq:thm_eq1} by $-\dfrac{\kappa}{2}$ for some $\kappa > 0$, then add to \eqref{eq:lem43}, we have
\begin{equation*}
\begin{array}{ll}
&\Exp{F(\theta_{t+1})} - \frac{\kappa}{2}\Exp{\norms{v_{t+1} - \nabla J(\theta_{t+1})}^2} \vspace{1ex}\\
 \ge & \Exp{F(\theta_t) } - \frac{(\kappa\beta^2+ \xi)}{2}\Exp{\norms{v_{t} - \nabla J(\theta_{t})}^2} + \frac{\eta^2 \alpha}{2} \Exp{\norms{\Gc_\eta(\theta_t)}^2} + \frac{1}{2}\left(\zeta - \frac{\kappa\beta^2\overline{C}\alpha^2}{B}\right)\Exp{\norms{\widehat{\theta}_{t+1} - \theta_t}^2} \vspace{1ex}\\
&- \frac{\kappa(1-\beta^2)\sigma^2}{2B} \vspace{1ex}\\
=& \Exp{F(\theta_t) } - \frac{\kappa}{2}\Exp{\norm{v_{t} - \nabla J(\theta_{t})}^2} + \frac{\eta^2 \alpha}{2} \Exp{\norms{\Gc_\eta(\theta_t)}^2}  - \frac{[\xi - \kappa(1-\beta^2)]}{2}\Exp{\norms{v_{t} - \nabla J(\theta_{t})}^2}  \vspace{1ex}\\
&+ \frac{1}{2}\left(\zeta - \frac{\kappa\beta^2\overline{C}\alpha^2}{B}\right)\Exp{\norms{\widehat{\theta}_{t+1} - \theta_t}^2} - \frac{\kappa(1-\beta^2)\sigma^2}{2B}.
\end{array}
\end{equation*}
Let us define $\overline{F}(\theta_t) := \Exp{F(\theta_{t})} - \frac{\kappa}{2}\Exp{\norms{v_{t} - \nabla J(\theta_{t+1})}^2}$.
Then, the last inequality can be written as
\begin{equation}\label{eq:thm_eq2}
\begin{array}{ll}
\overline{F}(\theta_{t+1}) &\ge \overline{F}(\theta_t) + \frac{\eta^2 \alpha}{2} \Exp{\norms{\Gc_\eta(\theta_t)}^2} - \frac{[\xi - \kappa(1-\beta^2)]}{2}\Exp{\norms{v_{t} - \nabla J(\theta_{t})}^2} \vspace{1ex}\\
& - \frac{\kappa(1-\beta^2)\sigma^2}{2B} + \frac{1}{2}\left(\zeta - \frac{\kappa\beta^2\overline{C}\alpha^2}{B}\right)\Exp{\norms{\widehat{\theta}_{t+1} - \theta_t}^2}.
\end{array}
\end{equation}
Suppose that $\eta$, $\alpha$, $\beta$ are chosen such that
\begin{equation}\label{eq:thm_eq3}
 \frac{2}{\eta} - L\alpha - 3 \ge \dfrac{\kappa \beta^2\overline{C}\alpha}{B} > 0~~~\text{and}~~~ \alpha(1+2\eta^2) \le \kappa(1 - \beta^2).
\end{equation}
Then, we have $\zeta \ge \dfrac{\kappa \beta^2\overline{C}\alpha^2}{B}$ and $\xi \le \kappa(1-\beta^2)$. 
By ignoring the non-negative terms in \eqref{eq:thm_eq2}, we can rewrite it as
\begin{equation*}
\overline{F}(\theta_{t+1}) \ge \overline{F}(\theta_t) + \dfrac{\eta^2 \alpha}{2} \Exp{\norm{\Gc_\eta(\theta_t)}^2}  - \dfrac{\kappa(1-\beta^2)\sigma^2}{2B}.
\end{equation*}
Summing the above inequality for $t = 0,\cdots, m$, we obtain
\begin{equation}\label{eq:thm_eq5}
\overline{F}(\theta_{m+1}) \ge \overline{F}(\theta_0) + \dfrac{\eta^2 \alpha}{2} \dsum_{t=0}^{m}\Exp{\norm{\Gc_\eta(\theta_t)}^2}  - \dfrac{\kappa(m+1)(1-\beta^2)\sigma^2}{2B}.
\end{equation}
Rearranging terms and multiply both sides by $\dfrac{2}{\eta^2\alpha}$, \eqref{eq:thm_eq5} becomes
\begin{equation}\label{eq:thm_eq6}
 \dsum_{t=0}^{m}\Exp{\norm{\Gc_\eta(\theta_t)}^2}  \le \dfrac{2}{\eta^2\alpha}\left[\overline{F}(\theta_{m+1})  - \overline{F}(\theta_0) \right] + \dfrac{\kappa(m+1)(1-\beta^2)\sigma^2}{\eta^2\alpha B}.
\end{equation}
Note that
\begin{equation*}
\overline{F}(\theta_0) = F(\theta_0) - \frac{\kappa}{2}\Exp{\norms{v_0 - \nabla J(\theta_0)}^2} \ge F(\theta_0) - \frac{\kappa\sigma^2}{2N},
\end{equation*}
and $\overline{F}(\theta_{m+1}) = F(\theta_{m+1}) - \frac{\kappa}{2}\Exp{\norms{v_{m+1} - \nabla J(\theta_{m+1})}^2} \le F(\theta_{m+1}) $. Using these estimate in \eqref{eq:thm_eq6}, we obtain
\begin{equation*}
\begin{array}{ll}
 \dsum_{t=0}^{m}\Exp{\norms{\Gc_\eta(\theta_t)}^2}  &\le \dfrac{2}{\eta^2\alpha}\left[ F(\theta_{m+1})  - F(\theta_0) \right] + \dfrac{\kappa\sigma^2}{\eta^2\alpha N} + \dfrac{\kappa(m+1)(1-\beta^2)\sigma^2}{\eta^2\alpha B} \vspace{1ex}\\
 &= \dfrac{2}{\eta^2\alpha}\left[ F(\theta_{m+1})  - F(\theta_0) \right] +  \dfrac{(m+1)\kappa\sigma^2}{\eta^2\alpha} \left[\dfrac{1}{N(m+1)} + \dfrac{(1-\beta^2)}{B} \right].
 \end{array}
\end{equation*}
Multiplying both sides by $\frac{1}{m+1}$, we have
\begin{equation}\label{eq:thm_eq7}
\frac{1}{m+1}\dsum_{t=0}^{m}\Exp{\norms{\Gc_\eta(\theta_t)}^2}  \le \frac{2}{\eta^2\alpha (m+1)}\left[ F(\theta_{m+1})  - F(\theta_0) \right]  +  \frac{\kappa\sigma^2}{\eta^2\alpha } \left[\frac{1}{N(m+1)} + \frac{(1-\beta^2)}{B} \right].
\end{equation}
Now we choose $\beta := 1 - \frac{\sqrt{B}}{\sqrt{N(m+1)}}$ so that the right-hand side of \eqref{eq:thm_eq7} is minimized. 
Note that if $1 \le B \le N(m+1)$, then $\beta \in [0,1)$. 

Let us choose $\eta := \frac{2}{4 + L\alpha} \le \frac{1}{2}$ which means $\zeta := \frac{2}{\eta} - L\alpha - 3 = 1$. 
We can satisfy the first condition of \eqref{eq:thm_eq3} by choosing $0 < \alpha \le \frac{B}{\kappa \overline{C}}$.

Besides, the second condition in \eqref{eq:thm_eq3} holds if $0 < \alpha \le \frac{\kappa(1-\beta^2)}{1+2\eta^2}$. 
Since we have $\eta \le \frac{1}{2}$ which leads to $1 + 2\eta^2 \le \frac{3}{2}$ and using $1-\beta^2 \ge 1-\beta = \frac{B^{1/2}}{N^{1/2}(m+1)^{1/2}}$ we derive the condition for $\alpha$ as
\begin{equation*}
0 < \alpha \le \frac{2\kappa \sqrt{B}}{3 \sqrt{N(m+1)}}.
\end{equation*}
Therefore, the overall condition for $\alpha$ is given as
\begin{equation*}
0 < \alpha \le \min\left\{1, \frac{B}{\kappa \overline{C}}, \frac{2\kappa \sqrt{B}}{3 \sqrt{N(m+1)}} \right\}.
\end{equation*}
If we choose $\kappa := \frac{\sqrt{3} [NB(m+1)]^{1/4} }{\sqrt{2 \overline{C}}}$, then we can update $\alpha$ as
\begin{equation}
\alpha := \frac{\hat{c} \sqrt{2}B^{3/4}}{\sqrt{3\overline{C}} [N(m+1)]^{1/4}}.
\end{equation}
Using $1\le B \le N(m+1)$, we can bound $\alpha \le \hat{c}\sqrt{ \frac{2B}{3\overline{C}}}$ then we can choose $\hat{c} \in \left(0,\sqrt{\frac{3\overline{C}}{2B}}\right]$ so that $\gamma \in (0,1]$.

With all the choices of $\beta$, $\eta$, $\alpha$, and $\kappa$ above, if we let the output $\tilde{\theta}_T$ be selected uniformly at random from $\left\{\theta_{t}\right\}_{t=0}^{m}$, then we have
\begin{equation}\label{eq:thm_eq8}
\begin{array}{ll}
\Exp{\norms{\Gc_\eta(\tilde{\theta}_T)}^2} &= \dfrac{1}{m+1} \dsum_{t=0}^{m}\Exp{\norm{\Gc_\eta(\theta_t)}^2}  \vspace{1ex}\\
&\le \dfrac{\sqrt{3\overline{C}}N^{1/4}}{\eta^2 \hat{c}\sqrt{2}[B(m+1)]^{3/4}}\left[ F(\theta_{m+1})  - F(\theta_0)\right] +  \dfrac{3 \sigma^2}{\eta^2 [BN(m+1)]^{1/2}}.
\end{array}
\end{equation}
Note that $\eta = \frac{2}{4 + L\alpha}$ and since $\alpha \le 1$ we have $\frac{1}{\eta^2} \le \frac{(4 + L)^2}{4}$. Plugging these into \eqref{eq:thm_eq8}, we obtain
\begin{equation}\label{eq:thm_eq9}
\begin{array}{ll}
\Exp{\norms{\Gc_\eta(\tilde{\theta}_T)}^2} &= \dfrac{1}{m+1} \dsum_{t=0}^{m}\Exp{\norm{\Gc_\eta(\theta_t)}^2}  \vspace{1ex}\\
&\le \dfrac{(4 + L)^2\sqrt{3\overline{C}}N^{1/4}}{4 \hat{c}\sqrt{2}[B(m+1)]^{3/4}}\left[ F(\theta_{m+1})  - F(\theta_0) \right] +  \dfrac{3(4 + L)^2 \sigma^2}{ 4[BN(m+1)]^{1/2}} \vspace{1ex}\\
&\le \dfrac{(4 + L)^2\sqrt{3\overline{C}}N^{1/4}}{4 \hat{c}\sqrt{2}[B(m+1)]^{3/4}}\left[ F^* - F(\theta_{0}) \right] +  \dfrac{3(4 + L)^2 \sigma^2}{ 4[BN(m+1)]^{1/2}},
\end{array}
\end{equation}
where we use the fact that $F(\theta_{m+1}) \le F^*$.
\Eproof

\subsection{Proof of Corollary~\ref{cor:complexity}: Trajectory Complexity Bound of Algorithm~\ref{alg:A1} and Algorithm~\ref{alg:A2}}\label{app:A4}
If we fix a batch size $B \in \mathbb{N}_+$ and choose $N := \tilde{c}\sigma^{8/3}\left[ B(m+1) \right]^{1/3}$ for some $\tilde{c} > 0$, \eqref{eq:thm_eq9} is equivalent to
\begin{equation*}\label{eq:cor1_eq1}
\begin{array}{ll}
\Exp{\norms{\Gc_\eta(\tilde{\theta}_T)}^2} &\le \dfrac{(4 + L)^2\sqrt{3\overline{C}}\tilde{c}^{1/4} \sigma^{2/3}}{4 \hat{c}\sqrt{2} [B(m+1)]^{2/3} }\left[ F^* - F(\overline{\theta}^{(0)}) \right] +  \dfrac{3(4 + L)^2 \sigma^{2/3}}{ 4 \tilde{c}^{1/2}[B(m+1)]^{2/3} }\vspace{1ex}\\
&= \left[ \dfrac{(4+L)^2\sqrt{3\overline{C}} \tilde{c}^{1/4}}{4\hat{c}\sqrt{2}}\left[F^* - F(\overline{\theta}^{(0)}) \right] + \dfrac{3(4+L)^2}{4\tilde{c}^{1/2}} \right] \frac{\sigma^{2/3}}{ [B(m+1)]^{2/3} } \vspace{1ex}\\
&=   \dfrac{\Psi_0\sigma^{2/3}}{ [ B(m+1)]^{2/3}},  
\end{array}
\end{equation*}
where we define 
\begin{equation}\label{eq:cor1_eq2}
\Psi_0 := \left[ \frac{(4+L)^2\sqrt{3\overline{C}} \tilde{c}^{1/4}}{4\hat{c}\sqrt{2}}\left[F^* - F(\overline{\theta}^{(0)}) \right] + \frac{3(4+L)^2}{4\tilde{c}^{1/2}} \right].
\end{equation}
Therefore, for any $\varepsilon > 0$, to guarantee $\Exp{\norms{\Gc_\eta(\tilde{\theta}_T)}^2} \le \varepsilon^2$, we need $\frac{\Psi_0\sigma^{2/3}}{ [B(m+1)]^{2/3}} = \varepsilon^2$ which leads to the total number of iterations
\begin{equation*}
T = m+1 = \frac{\Psi_0^{3/2}\sigma}{B\varepsilon^3} = \BigO{\dfrac{1}{\varepsilon^3}}.
\end{equation*}
The total number of proximal operations $\prox_{\eta Q}$ is also $\BigO{\frac{1}{\varepsilon^3}}$. In addition, the total number of trajectories is at most
\vspace{-2ex}
\begin{equation*}
\begin{array}{ll}
 N + 2B(m+1)&= \tilde{c}\sigma^{8/3}\left[ B(m+1) \right]^{1/3} + \dfrac{2\Psi_0\sigma}{\varepsilon^3} \vspace{1ex}\\
&= \tilde{c} \sigma^{8/3} \dfrac{\Psi_0^{1/3}\sigma{1/3}}{ \varepsilon}+ \dfrac{2\Psi_0\sigma}{\varepsilon^3} \vspace{1ex}\\
&= \BigO{ \dfrac{1}{\varepsilon} + \dfrac{1}{\varepsilon^3}} = \BigO{\dfrac{1}{\varepsilon^3}}.
\end{array}
\end{equation*}
This proves our the complexity of Algorithm~\ref{alg:A1}.

Next, let us denote the superscript $^{(s)}$ when the current stage is $s$ for $s = 0,\cdots, S-1$. Note that from the first inequality of \eqref{eq:thm_eq9}, for any stage $s = 0,\dots,S-1$, the following holds
\begin{equation*}\label{eq:cor2_eq1}
\begin{array}{ll}
\dfrac{1}{m+1} \dsum_{t=0}^{m}\Exp{\norms{\Gc_\eta(\theta_t^{(s)})}^2}&\le \dfrac{(4 + L)^2\sqrt{3\overline{C}}N^{1/4}}{4 \hat{c}\sqrt{2}[B(m+1)]^{3/4}}\left[ F(\theta_{m+1}^{(s)})  - F(\theta_0^{(s)})  \right] +  \dfrac{3(4 + L)^2 \sigma^2}{ 4[BN(m+1)]^{1/2}}.
\end{array}
\end{equation*}
Summing for $s = 0,\cdots,S-1$ and multiply both sides by $\dfrac{1}{S}$ yields
\begin{equation}\label{eq:cor2_eq2}
{\!\!\!\!\!\!\!\!}\begin{array}{ll}
\frac{1}{S(m+1)}\dsum_{s=0}^{S-1} \dsum_{t=0}^{m}\Exp{\norms{\Gc_\eta(\theta_t^{(s)})}^2} {\!\!\!\!\!}&\le \frac{(4 + L)^2\sqrt{3\overline{C}}N^{1/4}}{4 \hat{c}\sqrt{2}[B(m+1)]^{3/4}S}\left[ F(\theta_{m+1}^{(S-1)})  - F(\theta_0^{(0)})  \right] +  \frac{3(4 + L)^2 \sigma^2}{ 4[BN(m+1)]^{1/2}S} \vspace{1ex}\\
&\le \frac{(4 + L)^2\sqrt{3\overline{C}}N^{1/4}}{4 \hat{c}\sqrt{2}[B(m+1)]^{3/4}S}\left[ F^* - F(\theta_{0}^{(0)})  \right] +  \frac{3(4 + L)^2 \sigma^2}{ 4[BN(m+1)]^{1/2}S},
\end{array}{\!\!\!\!\!}
\end{equation}
where we use $F(\theta_{m+1}^{(S-1)}) \le F^*$ again.

If we also fix a batch size $B \in \mathbb{N}_+$ and choose $N := \tilde{c}\sigma^{8/3}\left[ B(m+1) \right]^{1/3}$ for some $\tilde{c} > 0$, and select $\tilde{\theta}_T$ uniformly random from $\{\theta_t^{(s)}\}_{t=0,\cdots,m}^{s=1,\cdots,S}$, then, similar to \eqref{eq:cor1_eq1}, \eqref{eq:cor2_eq2} can be written as
\begin{equation*}\label{eq:cor2_eq3}
\begin{array}{ll}
\Exp{\norms{\Gc_\eta(\tilde{\theta}_T)}^2} &= \dfrac{1}{S(m+1)}\dsum_{s=0}^{S-1} \dsum_{t=0}^{m}\Exp{\norms{\Gc_\eta(\theta_t^{(s)})}^2} \vspace{1ex}\\
&\le \dfrac{(4 + L)^2\sqrt{3\overline{C}}\tilde{c}^{1/4} \sigma^{2/3}}{4 \hat{c}\sqrt{2} [B(m+1)]^{2/3} S}\left[ F^* - F(\theta_{0}^{(0)}) \right] +  \dfrac{3(4 + L)^2 \sigma^{2/3}}{ 4 \tilde{c}^{1/2}[B(m+1)]^{2/3} S} \vspace{1ex}\\
&= \left[ \dfrac{(4+L)^2\sqrt{3\overline{C}} \tilde{c}^{1/4}}{4\hat{c}\sqrt{2}}\left[F^* - F(\theta_{0}^{(0)}) \right] + \dfrac{3(4+L)^2}{4\tilde{c}^{1/2}} \right] \dfrac{\sigma^{2/3}}{ [B(m+1)]^{2/3} S} \vspace{1ex}\\
&\le   \dfrac{\Psi_0\sigma^{2/3}}{ [ SB(m+1)]^{2/3}}, \vspace{1ex}\\
\end{array}
\end{equation*}
where we use $\Psi_0$ defined in \eqref{eq:cor1_eq2} and $\frac{1}{S} \le \frac{1}{S^{2/3}}$ for any $S \ge 1$.

Therefore, to guarantee $\Exp{\norms{\Gc_\eta(\tilde{\theta}_T)}^2} \le \varepsilon^2$ for any $\varepsilon > 0$, we need $\frac{\Psi_0\sigma^{2/3}}{ [SB(m+1)]^{2/3}} = \varepsilon^2$ which leads to the total number of iterations
\begin{equation*}
T = S(m+1) = \frac{\Psi_0^{3/2}\sigma}{B\varepsilon^3} = \BigO{\frac{1}{\varepsilon^3}}.
\end{equation*}
The total number of proximal operations $\prox_{\eta Q}$ is also $\BigO{\frac{1}{\varepsilon^3}}$. In addition, the total number of trajectories is at most
\vspace{-2ex}
\begin{equation*} 
\begin{array}{ll}
 S\left[N + 2B(m+1)\right]&= S\left[\tilde{c}\sigma^{8/3}\left[ B(m+1) \right]^{1/3} + \dfrac{2\Psi_0\sigma}{\varepsilon^3}\right] \vspace{1ex}\\
&= S\left[\tilde{c} \sigma^{8/3} \dfrac{\Psi_0^{1/3}\sigma{1/3}}{ \varepsilon}+ \dfrac{2\Psi_0\sigma}{\varepsilon^3}\right] \vspace{1ex}\\
&= \BigO{ \dfrac{1}{\varepsilon} + \dfrac{1}{\varepsilon^3}} = \BigO{\dfrac{1}{\varepsilon^3}},~\text{for any $S \ge 1$.}
\end{array}
\end{equation*}
Hence, we obtain the conclusion of Corollary \ref{cor:complexity}.
\Eproof

\section{Configurations of Algorithms in Section~\ref{sec:experiments}}\label{sec:exp_conf}
Let us describe in detail the configuration of our experiments in Section~\ref{sec:experiments}.
We set $\beta := 0.99$ for HSPGA and $\alpha := 0.99$ for ProxHSPGA in all experiments. 
To choose the learning rate, we conduct a grid search over different choices. For \texttt{Acrobot-v1}, \texttt{Cart pole-v0}, and \texttt{Mountain Car-v0} environments, we use the grid containing $\{0.0005,0.001,0.0025,0.005,0.0075,0.01 \}$. 
Meanwhile, we use $\{0.0005,0.00075,0.001,0.0025,0.005\}$ for the remaining environments. 
The snapshot batch-sizes are also chosen from $\{10,25,50,100 \}$ while the mini-batch sizes are selected from $\{3,5,10,15,20,25 \}$. 
More details about the selected parameters for each experiment are shown in Table~\ref{tab:config}.

\renewcommand{\arraystretch}{1.2}
\begin{table}[]
\caption{The configuration of different algorithms on discrete and continuous control environments}
\label{tab:config}
\vspace{2ex}
\resizebox{\textwidth}{!}{%
\begin{tabular}{|l|l|c|c|c|c|c|c|c|}
\hline
\multirow{2}{*}{Environment}  & \multirow{2}{*}{Algorithm} & Policy & Discount &Trajectory & Minibatch & Snapshot & Learning & Epoch \\ 
& & Network & Factor $\gamma$ & Length H & Size & Batchsize & Rate & Length $m$\\
\hline
\multirow{3}{*}{CartPole-v0} & GPOMDP & \multirow{3}{*}{$4\times 8\times 2$} & \multirow{3}{*}{0.99} & \multirow{3}{*}{200} & 10 &  & $10^{-3}$ &  \\ \cline{2-2} \cline{6-9} 
 & SVRPG &  &  &  & 10 & 25 & $5\times 10^{-3}$ & 3 \\ \cline{2-2} \cline{6-9} 
 & HSPGA &  &  &  & 5 & 25 & $5\times 10^{-3}$ & 3 \\ \hline
\multirow{3}{*}{Acrobot-v1} & GPOMDP & \multirow{3}{*}{$6\times16\times3$} & \multirow{3}{*}{0.999} & \multirow{3}{*}{500} & 10 &  & $2.5\times 10^{-3}$ &  \\ \cline{2-2} \cline{6-9} 
 & SVRPG &  &  &  & 5 & 10 & $5\times 10^{-3}$ & 3 \\ \cline{2-2} \cline{6-9} 
 & HSPGA &  &  &  & 3 & 10 & $5\times 10^{-3}$ & 3 \\ \hline
\multirow{3}{*}{MoutainCar-v0} & GPOMDP & \multirow{3}{*}{$2\times8\times1$} & \multirow{3}{*}{0.999} & \multirow{3}{*}{1000} & 25 &  & $5\times 10^{-3}$ &  \\ \cline{2-2} \cline{6-9} 
 & SVRPG &  &  &  & 10 & 50 & $7.5\times 10^{-3}$ & 3 \\ \cline{2-2} \cline{6-9} 
 & HSPGA &  &  &  & 5 & 50 & $7.5\times 10^{-3}$ & 3 \\ \hline
\multirow{4}{*}{RoboschoolInvertedPendulum-v1} & GPOMDP & \multirow{4}{*}{$5\times16\times1$} & \multirow{4}{*}{0.999} & \multirow{4}{*}{1000} & 20 &  & $7.5\times 10^{-4}$ &  \\ \cline{2-2} \cline{6-9} 
 & SVRPG &  &  &  & 10 & 50 & $10^{-3}$ & 3 \\ \cline{2-2} \cline{6-9} 
 & HSPGA &  &  &  & 5 & 50 & $10^{-3}$ & 3 \\ \cline{2-2} \cline{6-9} 
 & ProxHSPGA &  &  &  & 5 & 50 & $10^{-3}$ & 3 \\ \hline
\multirow{4}{*}{Swimmer-v2} & GPOMDP & \multirow{4}{*}{$8\times 32\times 32\times 2$} & \multirow{4}{*}{0.99} & \multirow{4}{*}{500} & 50 &  & $5\times 10^{-4}$ &  \\ \cline{2-2} \cline{6-9} 
 & SVRPG &  &  &  & 5 & 50 & $5\times 10^{-4}$ & 3 \\ \cline{2-2} \cline{6-9} 
 & HSPGA &  &  &  & 5 & 50 & $5\times 10^{-4}$ & 3 \\ \cline{2-2} \cline{6-9} 
 & ProxHSPGA &  &  &  & 5 & 50 & $5\times 10^{-4}$ & 3 \\ \hline
\multirow{4}{*}{Hopper-v2} & GPOMDP & \multirow{4}{*}{$11\times 32\times 32\times 3$} & \multirow{4}{*}{0.99} & \multirow{4}{*}{500} & 50 &  & $5\times 10^{-4}$ &  \\ \cline{2-2} \cline{6-9} 
 & SVRPG &  &  &  & 5 & 50 & $5\times 10^{-4}$ & 3 \\ \cline{2-2} \cline{6-9} 
 & HSPGA &  &  &  & 5 & 50 & $5\times 10^{-4}$ & 3 \\ \cline{2-2} \cline{6-9} 
 & ProxHSPGA &  &  &  & 5 & 50 & $5\times 10^{-4}$ & 3 \\ \hline
 \multirow{4}{*}{Walker2d-v2} & GPOMDP & \multirow{4}{*}{$17\times 32\times 32\times 6$} & \multirow{4}{*}{0.99} & \multirow{4}{*}{500} & 50 &  & $5\times 10^{-4}$ &  \\ \cline{2-2} \cline{6-9} 
 & SVRPG &  &  &  & 5 & 50 & $5\times 10^{-4}$ & 3 \\ \cline{2-2} \cline{6-9} 
 & HSPGA &  &  &  & 5 & 50 & $5\times 10^{-4}$ & 3 \\ \cline{2-2} \cline{6-9} 
 & ProxHSPGA &  &  &  & 5 & 50 & $5\times 10^{-4}$ & 3 \\ \hline
\end{tabular}%
}
\end{table}
\renewcommand{\arraystretch}{1}

\section{Additional Numerical Results}\label{app:num_exp}
\begin{figure}[htp!]
\begin{center}
\includegraphics[width = 0.32\textwidth]{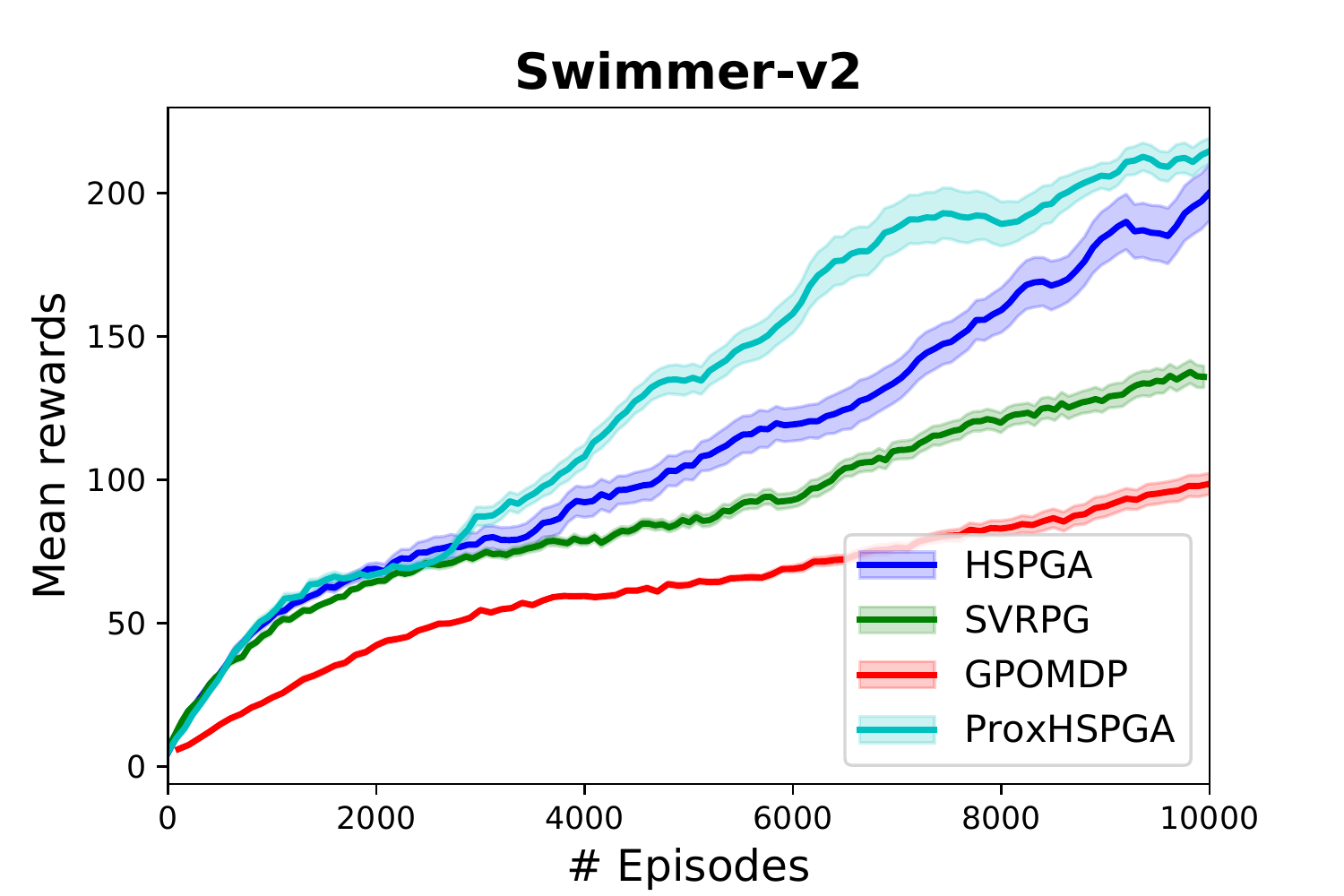}
\includegraphics[width = 0.32\textwidth]{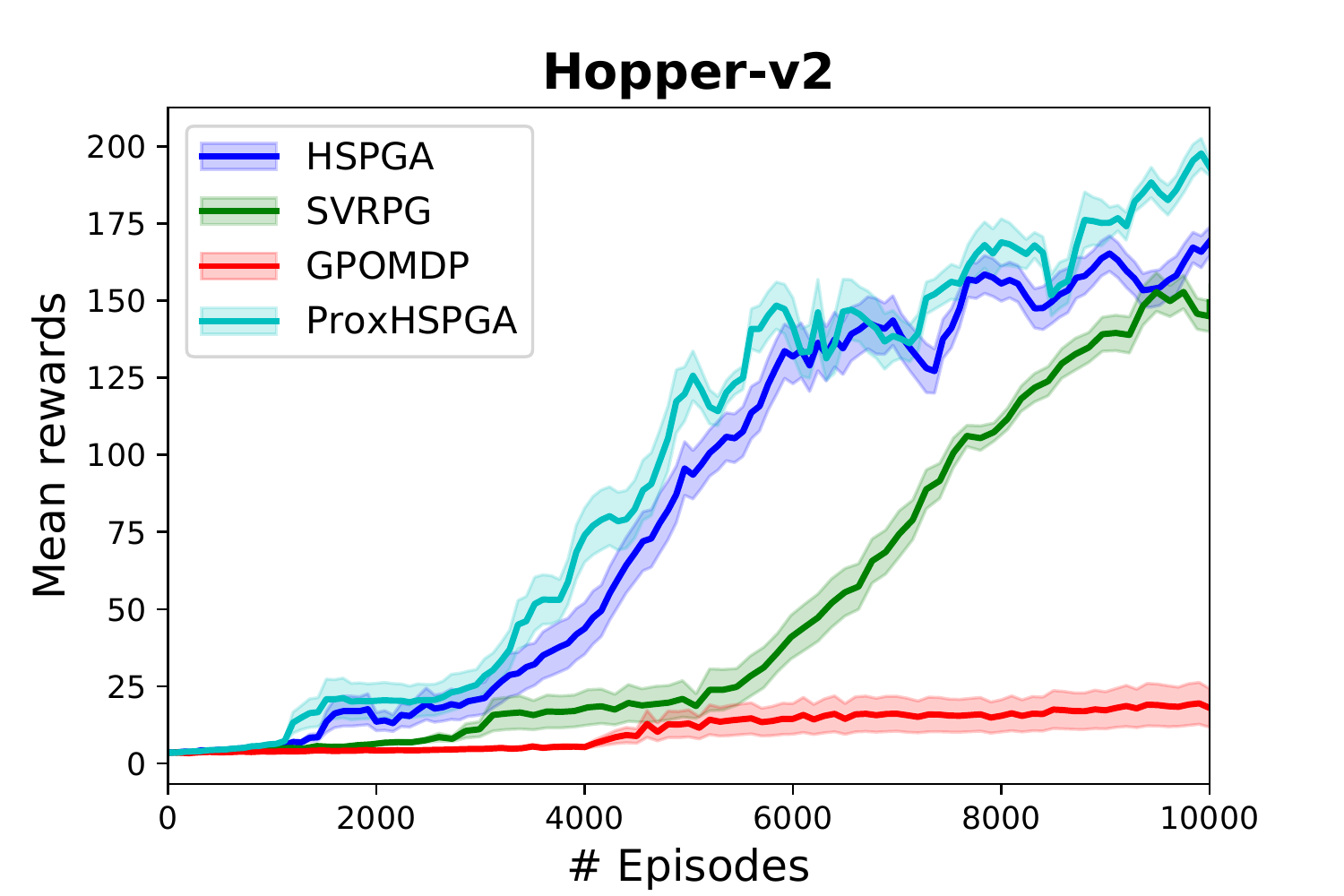}
\includegraphics[width = 0.32\textwidth]{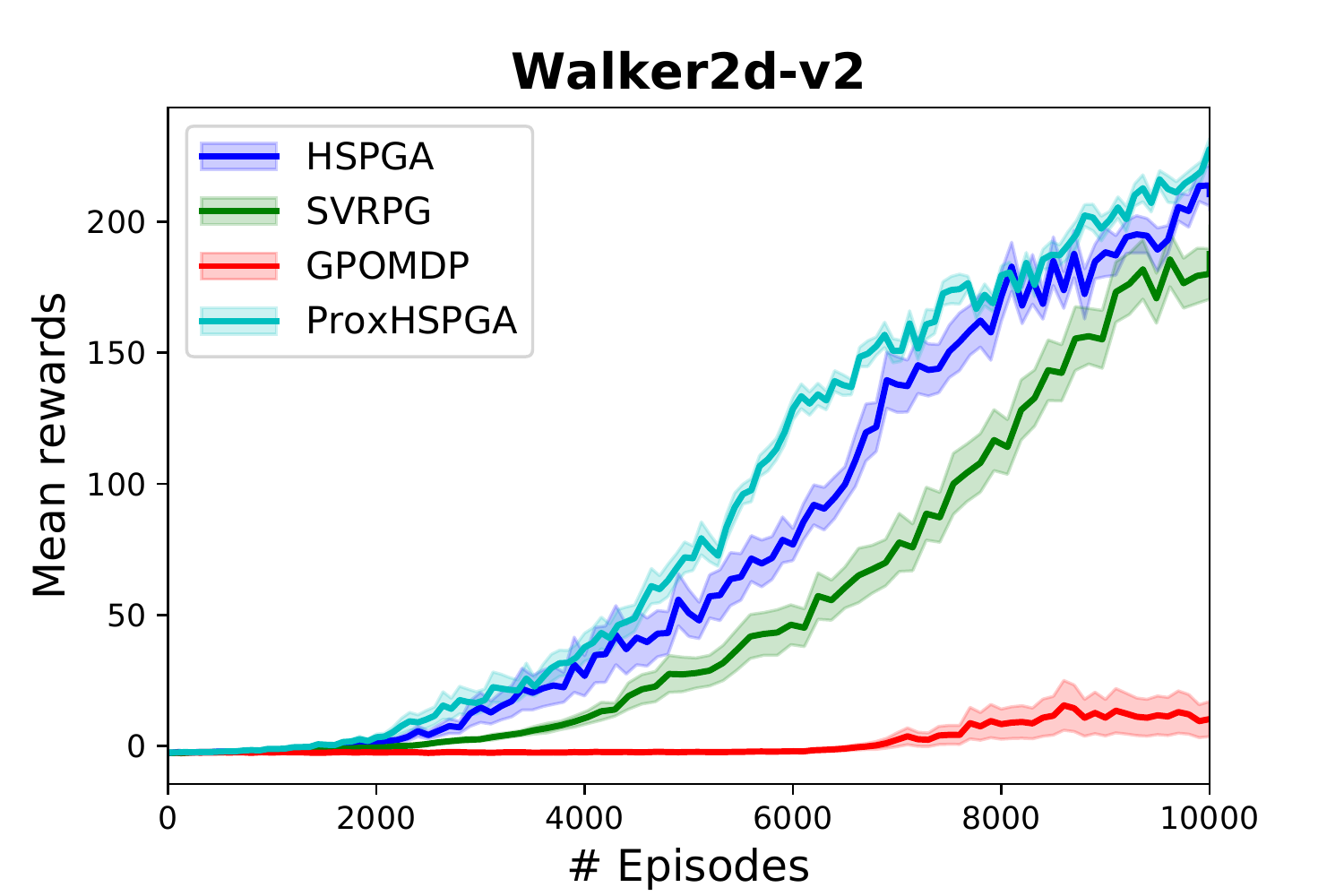}
\caption{The performance of 4 algorithms on the composite vs. the non-composite settings using several \texttt{Mujoco} environments.}\label{fig:mujoco}
\end{center}
\end{figure}

Due to space limit in the main text, we show here another evidence on the effect of regularizers to policy optimization problems by carrying out an additional example on other continuous control tasks in \texttt{Mujoco}.
The results are presented in Figure~\ref{fig:mujoco}.

Again, Figure~\ref{fig:mujoco} still reveals the benefit of adding a regularizer, which potentially gains more reward than without using regularizer. 
We believe that the choice of regularizer is also critical and may lead to different performance.
We refer to \citep{liu2019regularization} for more evidence of using regularizers in reinforcement learning. 

\end{document}